\documentclass[journal]{IEEEtran}
\usepackage{amsmath,amsfonts,amssymb}
\usepackage{algorithmic}
\usepackage{algorithm}
\usepackage{array}
\usepackage[caption=false,font=normalsize,labelfont=sf,textfont=sf]{subfig}
\usepackage{textcomp}
\usepackage{stfloats}
\usepackage{url}
\usepackage{verbatim}
\usepackage{graphicx}
\usepackage{cite}
\usepackage{xspace}
\usepackage{xcolor}
\usepackage{bbding}
\usepackage{lineno}
\usepackage{booktabs}
\usepackage{multirow}
\usepackage{arydshln}
\usepackage{enumitem}

\usepackage{color}

\newcommand{\cq}{\textcolor{black}}

\begin{document}

\title{SDPL: Shifting-Dense Partition Learning for UAV-view Geo-localization}

\author{Quan Chen, Tingyu Wang, Zihao Yang, Haoran Li, Rongfeng Lu, Yaoqi Sun\\
Bolun Zheng and Chenggang Yan
\thanks{Copyright © 2024 IEEE. Personal use of this material is permitted. However, permission to use this material for any other purposes must be obtained from the IEEE by sending an email to pubs-permissions@ieee.org.}
\thanks{This work was supported in part by the Key R\&D Program of Zhejiang under Grant No. 2023C01044; in part by the National Nature Science Foundation of China No. 62371175; and in part by the Fundamental Research Funds for the Provincial Universities of Zhejiang under Grants No. GK239909299001-013.}
\thanks{Quan Chen, Zihao Yang, Haoran Li, Rongfeng Lu, Bolun Zheng and Chenggang Yan are with the School of Automation, Hangzhou Dianzi University, Hangzhou 310018, China (e-mail:chenquan@hdu.edu.cn; zihaoyang@hdu.edu.cn; lhr970315@gmail.com; rongfeng-lu@hdu.edu.cn; blzheng@hdu.edu.cn; cgyan@hdu.edu.cn).}
\thanks{Tingyu Wang and Yaoqi Sun are with the School of Communication
Engineering, Hangzhou Dianzi University, Hangzhou 310018, China, and also with the Lishui Institute of Hangzhou Dianzi University, Lishui 323000, China (e-mail:wongtyu@hdu.edu.cn; syq@hdu.edu.cn).}
\thanks{Tingyu Wang and Yaoqi Sun are the Corresponding Authors.}
}

\markboth{Journal of \LaTeX\ Class Files,~Vol.~14, No.~8, August~2021}%
{Shell \MakeLowercase{\textit{et al.}}: A Sample Article Using IEEEtran.cls for IEEE Journals}


\maketitle

\begin{abstract}
Cross-view geo-localization aims to match images of the same target from different platforms,~\textit{e.g.}, drone and satellite.
It is a challenging task due to the changing appearance of targets and environmental content from different views.
Most methods focus on obtaining more comprehensive information through feature map segmentation, while inevitably destroying the image structure, and are sensitive to the shifting and scale of the target in the query.
\cq{To address the above issues, we introduce simple yet effective part-based representation learning, shifting-dense partition learning~(SDPL).
We propose a dense partition strategy~(DPS), dividing the image into multiple parts to explore contextual information while explicitly maintaining the global structure.
To handle scenarios with non-centered targets, we further propose the shifting-fusion strategy, which generates multiple sets of parts in parallel based on various segmentation centers, and then adaptively fuses all features to integrate their anti-offset ability.
Extensive experiments show that SDPL is robust to position shifting, and performs competitively on two prevailing benchmarks, University-1652 and SUES-200.
In addition, SDPL shows satisfactory compatibility with a variety of backbone networks (\textit{e.g.}, ResNet and Swin).}
\url{https://github.com/C-water/SDPL_release.}
\end{abstract}


\section{Introduction}
\IEEEPARstart{U}{nmanned}  Aerial Vehicles~(UAVs) have gained popularity due to their ability to capture high-quality multimedia data. 
As an emerging vision platform, UAVs capture images with a larger field of view, freer angles, and less occlusion than ground-view images. 
Benefiting from excellent target visibility, UAVs have been widely deployed in many fields such as accuracy delivery~\cite{dissanayaka2023review,sorbelli2023wind}, automatic driving~\cite{khan2017uav,wang2019development} and agriculture and plant protection~\cite{deng2018uav,huang2022evaluation}.
Geo-localization is a primary task underpinning applications of UAVs.
The current positioning and navigation of UAVs mainly rely on GPS~\cite{rieke2012high,zimmermann2017precise}, which requires a high-quality communication environment. 
When the GPS signal is weak or denied, the positioning system will incorrectly determine the geographic location of UAVs. 
Therefore, image-based cross-view geo-localization assisting GPS to achieve accurate positioning develops into a meaningful hot research.
Cross-view image matching technology is the key to solve this problem, \textit{i.e.}, the geographic positions of images can be determined by matching drone-view and satellite-view images, which faces challenges due to the vast variations in visual appearance.
Most recent learning-based cross-view geo-localization methods~\cite{zheng2020university,zhu2023uav,wang2021each,zheng2023uavm,lin2022joint,sun2023f3,ding2020practical,wang2022multiple,zhao2024transfg} attempt to learn a mapping function that projects multi-view images into one shared semantic space for feature discrimination, \textit{i.e.}, images with the same location are close, and vice versa.
To fully mine the contextual information around the geographic target, several part-based representation learning strategies~\cite{wang2021each,dai2021transformer,zhuang2021faster,li2023drone} have been introduced, which inevitably damage the image structure and ignore the influence of position deviations.

\begin{figure}[!t]
    \centering
    \includegraphics[width=1.0\linewidth]{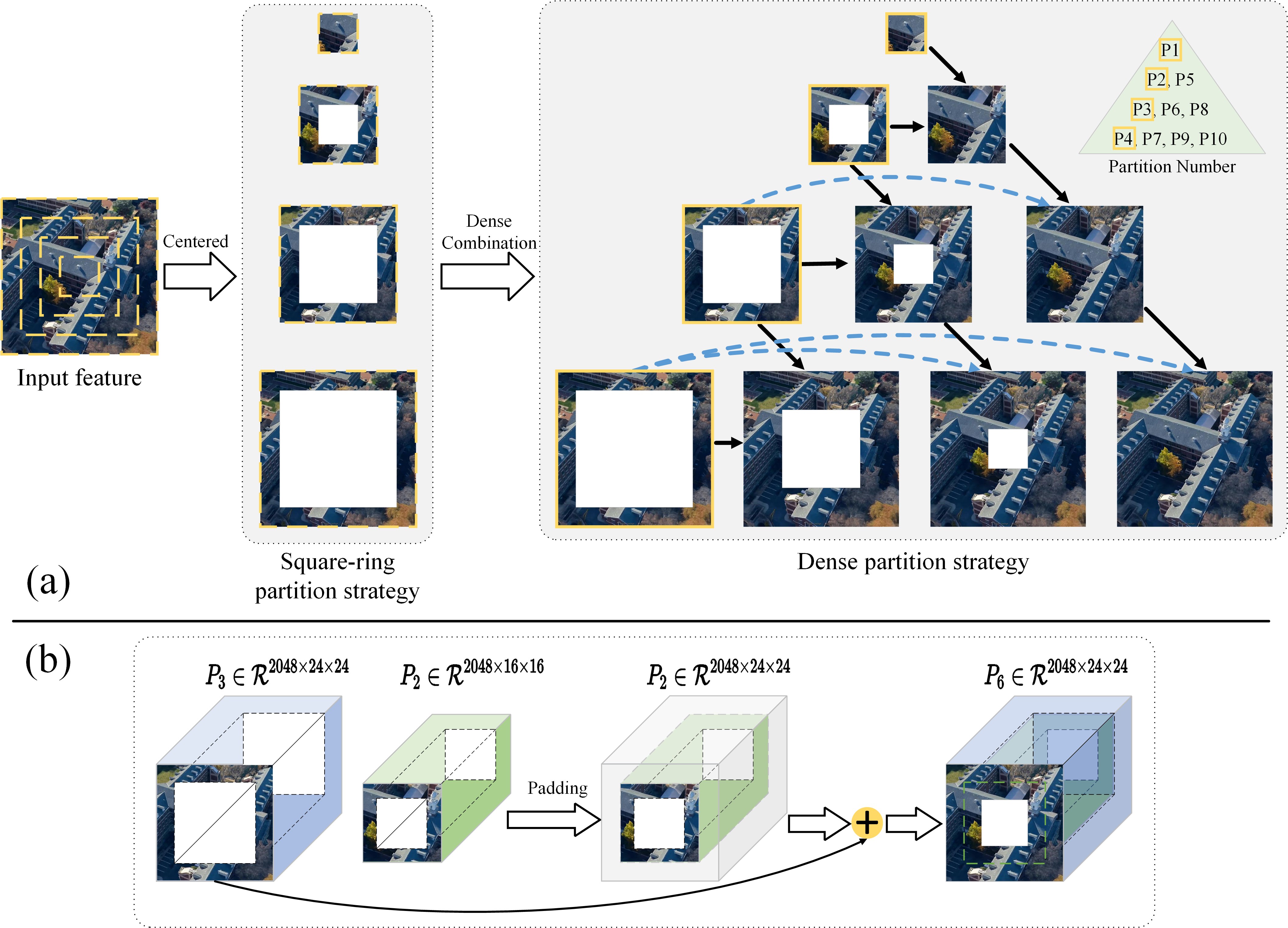}
    \caption{Dense partition strategy. (a) Yellow dashed boxes: base parts generated by LPN~\cite{wang2021each}. DPS recombines   base parts to mine local context information while preserving global structure. Part number $P$ is marked in green triangle; (b) Recombination process for parts with various resolutions. The low-resolution features are padded to ensure resolution consistency of blocks.}
    \label{denselpn img}
\end{figure}

\begin{figure}[!t]
    \centering
    \includegraphics[width=1.0\linewidth]{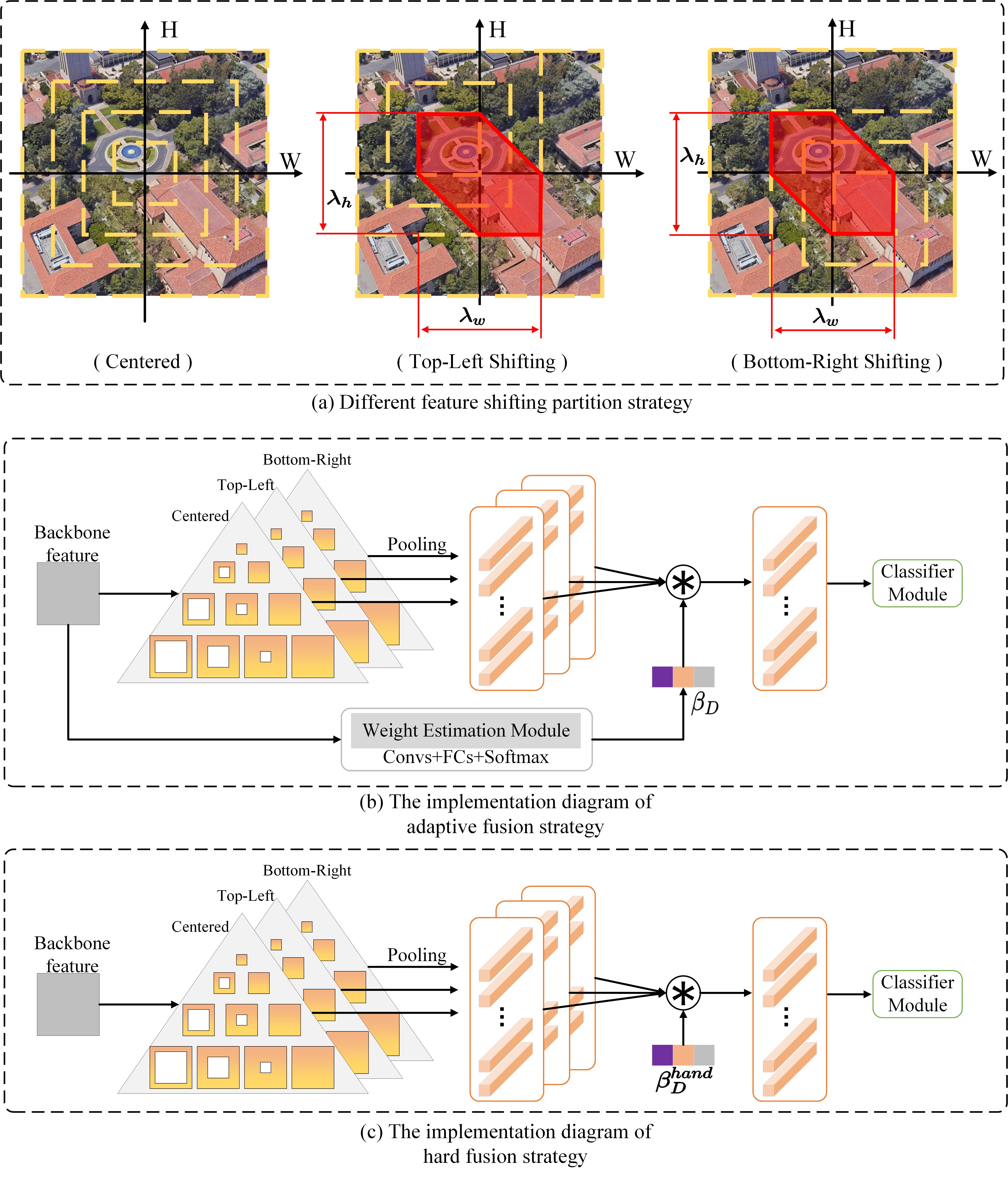}
    \caption{Feature shifting-fusion strategy. (a)  Examples of partitions generated by DPS with various segmentation centers: centered, top-left shifting, bottom-right shifting; (b)  Adaptive fusion strategy fuses three sets of parts using   parameter $\beta_{D}$ learned from weight estimation module; (c) Hyper-parameter $\beta_{D}^{hand}$ is adopted to fuse three sets of parts for ablation experiments.}
    \label{shifting strategy img}
\end{figure}

\begin{figure*}[!t]
\centering
\includegraphics[width=1.0\linewidth]{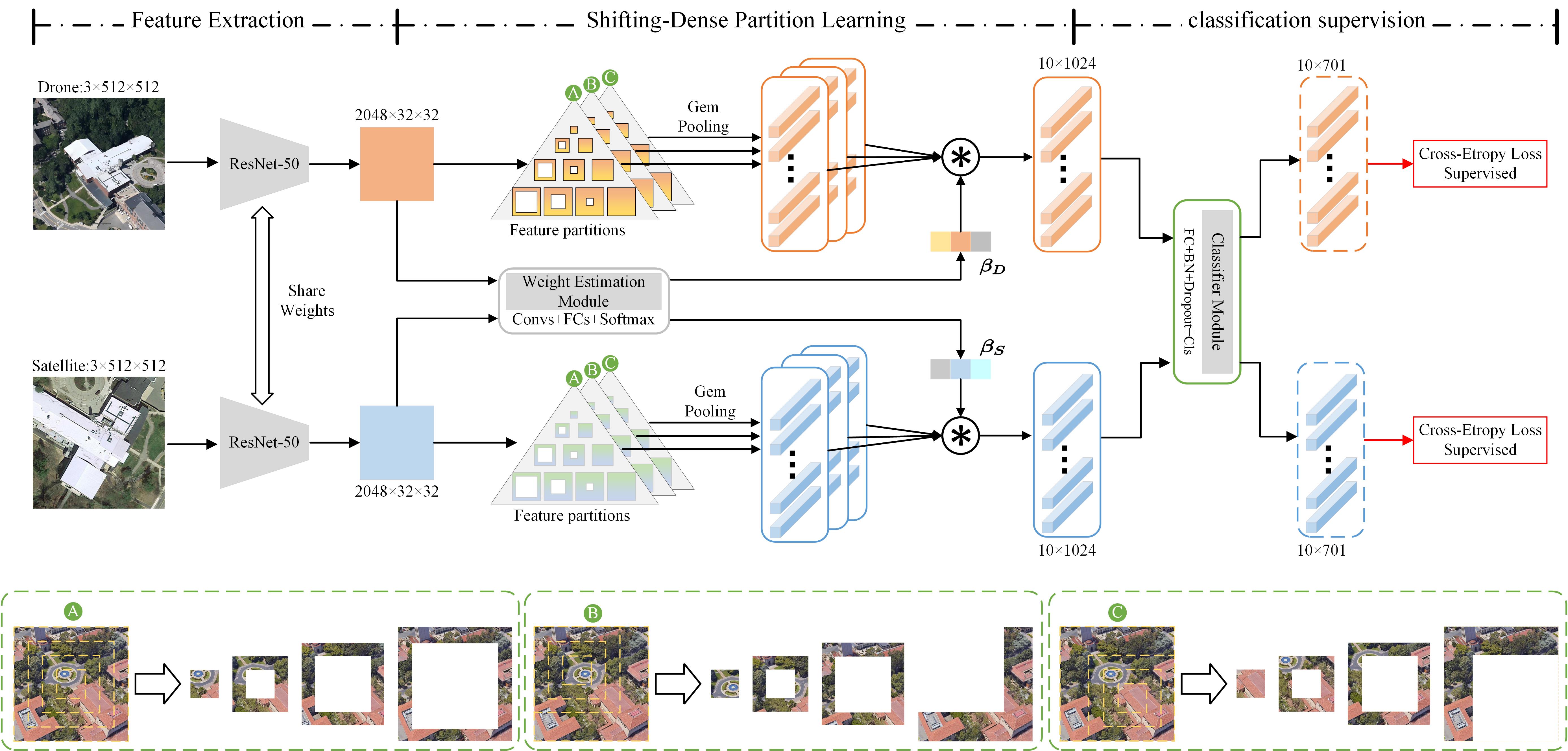}
\caption{Overview of SDPL framework.
A--C (bottom): Diagrams of dense partition strategy with various segmentation centers.
During testing, part-level image representation is extracted before  classification layer in classifier module, and measures similarity by Euclidean distance.}
\label{model}
\end{figure*}

To address the above issues, we propose shifting-dense partition learning~(SDPL), consisting of a dense partition strategy~(DPS) and shifting-fusion strategy~(SFS).
As shown in Fig.~\ref{denselpn img}, compared with the representative square-ring partition strategy~\cite{wang2021each}, DPS divides features into shape-diverse parts to mine more fine-grained representation, and incrementally dilates the boundaries of partitions to mitigate local structural damage.
\cq{Meanwhile, DPS explicitly preserves global features to increase the robustness of SDPL to scale changes and target offsets.}
DPS is based on the imperfect assumption that the target is located in the central region.
\cq{To enhance the robustness of DPS in non-central target scenarios, we further introduce the shifting-fusion strategy, which consists of diagonal-shifting partition and adaptive feature fusion~(Figs.~\ref{shifting strategy img}(a) and (b)).
The first step generates multiple sets of part-level features based on DPS with various segmentation centers, which are offset diagonally along the image, in turn middle, top-left and bottom-right.
Adjusting the segmentation center can enhance the anti-offset ability of DPS in a specific direction, but the scheme of stacking DPS is inefficient.
To overcome this limitation, we introduce an adaptive fusion strategy, which leverages a weight estimation module to adaptively fuse multiple sets of partitions, thus obtaining more robust anti-offset potential at a low cost.}
Note that the proposed SDPL includes phases of feature extraction, shifting-dense partition learning, and classification supervision~(Fig.~\ref{model}).

Our main contributions are as follows:
\begin{itemize}
    \item We propose SDPL, including dense partition and shifting-fusion strategies, to achieve accurate cross-view geo-localization against position deviations and scale changes;
    \item To mitigate the impact of scale changes and target offsets, we propose DPS to divide features into shape-diverse parts, thus mining fine-grained representation while preserving global structure;
    \item \cq{For degradation caused by position offset, we design a shifting-fusion strategy, which adaptively fuses multiple sets of partitions with various segmentation centers to  integrate their anti-offset ability;}
    \item \cq{Extensive experiments show that our SDPL achievescompetitive retrieval accuracy on two public datasets, \textit{i.e.}, University-1652~\cite{zheng2020university} and SUES-200~\cite{zhu2023sues}. Ablation experiments show that SDPL has superior anti-offset property.}
\end{itemize}

The rest of this article is organized as follows. 
Section~\ref{Related Works} presents related work. Section~\ref{Methods} introduces the proposed methods. Section~\ref{Experiments and Results} presents our experimental results, and Section~\ref{Conclusions} relates our conclusions.

\section{Related Work}\label{Related Works}
We briefly review related work, including deep cross-view geo-localization and part-based representation learning.

\subsection{Cross-view Geo-localization}
\cq{Benefiting from  complementary content, multi-view matching technology has made great breakthroughs in plentiful visual tasks, such as person re-identification~\cite{liu2022data,ji2022asymmetric} and image template matching~\cite{li2023image,ye2022multiscale}.
A dominant pipeline maps images into the shared semantic space and constrains matched features to have similar distributions.
Several studies~\cite{liu2019stochastic,workman2015wide} have focused on metric learning to optimize the distance between image pairs. 
Li~\textit{et al.}~\cite{li2023image} performed dense and consistent InfoNCEloss during matching,  encouraging correspondence matching across modalities at the pixel level.
Compared with the aforementioned vision-based matching tasks~(matching images from the similar platform), the challenge of   cross-view geo-localization is that images acquired from different platforms~(\textit{e.g.}, ground- and satellite-view) have a distinct visual appearance.}

Cross-view geo-localization focuses on the matching of ground and satellite views and of drone and satellite views.
Limited by data acquisition, early works~\cite{castaldo2015semantic,lin2013cross,senlet2011framework} attempted to extract handcrafted features to address this task.
Benefiting from the success of deep convolutional neural networks~(CNNs), recent geo-localization works~\cite{workman2015location,workman2015wide,lin2015learning,zhai2017predicting,liu2019lending,li2023patch,sun2024tirsa,li2024unleashing} have focused on learning deep representations, 
and constructed several public datasets containing ground-to-satellite image pairs,~\textit{e.g.}, CVUSA~\cite{zhai2017predicting} and CVACT~\cite{liu2019lending}.
Workman~\textit{et al.}~\cite{workman2015location} first utilized a pretrained CNN to extract high-level features for the cross-view localization task, and proved that the features contain semantic information about the geographic location.
Inspired by the use of Siamese networks~\cite{chopra2005learning} in learning viewpoint-invariant features, Zhai~\textit{et al.}~\cite{zhai2017predicting} inserted   NetVLAD~\cite{arandjelovic2016netvlad} into a Siamese-like network, making image descriptors robust against large viewpoint changes.
Shi~\textit{et al.}~\cite{shi2019spatial} explored domain alignment, and applied a polar transform to warp aerial images for multi-view alignment.
\cq{Considering the gain of hard negative examples to model performance, the GPS-sampling and dynamic similarity sampling strategies were proposed for mining hard negatives~\cite{deuser2023sample4geo}.}

With the popularity of UAV devices, several drone-based datasets have been proposed, promoting the development of UAV-view geo-localization.
Zheng~\textit{et al.}~\cite{zheng2020university} presented the University-1652 dataset, composed of drone-satellite image pairs, and examined image-retrieval tasks from a classification perspective.
Furthermore, Zheng~\textit{et al.}~\cite{zheng2023uavm} extended the University-1652 with extra satellite-view gallery distractors to build University-160k.
Zhu~\textit{et al.}~\cite{zhu2023sues} constructed SUES-200, focusing on the effect of drone-views at different heights on geo-localization.
\cq{DenseUAV~\cite{dai2023vision}, as the first low-altitude urban scene dataset, was presented for the UAV self-positioning task.}
Stronger backbone networks can significantly improve the accuracy of image matching. 
Dai~\textit{et al.}~\cite{dai2021transformer} adopted ViT~\cite{dosovitskiy2020image} as a backbone, and achieved competitive performance with CNN-based frameworks.
To fully mine the context information of aerial-view images, Wang~\textit{et al.}~\cite{wang2021each} designed a rotation-invariant square-ring partition strategy to explicitly explore contextual information.
Shen~\textit{et al.}~\cite{shen2023mccg} designed an effective multiple-classifier structure to capture rich discriminative features.
\cq{Different from existing works, the proposed method focuses on mitigating performance degradation caused by position deviations.}

\subsection{Part-based Representation Learning}
Local features guide models to learn more comprehensive features, which has been proven effective in many fields such as object detection~\cite{leibe2008robust,weber2000towards} and recognition~\cite{amit2007pop,crandall2005spatial,fergus2003object}.
Ojala~\textit{et al.}~\cite{ojala2002multiresolution} proposed a local binary pattern descriptor to extract rotation-invariant features.
Lowe~\cite{lowe1999object} proposed a scale-invariant feature transform descriptor for the image matching task, which utilized the classic difference of the Gaussian pyramid to summarize descriptions of local image structures from local neighborhoods around each interest point.
In the spirit of the conventional part-based descriptor, recent works~\cite{qian2020stripe,sun2019dissecting,zhong2019invariance,song2019generalizable,tian2021uav,zheng2019implicit} have adopted CNN-based models to acquire local feature representations.
Spindle-Net~\cite{zhao2017spindle} leveraged body joints to capture semantic features from different body regions, enabling the alignment of macro- and micro-body features across images.
\cq{ACSA~\cite{ji2022asymmetric} employed a global-level alignment module to align image and text on a global scale, with an asymmetric cross-attention module to dynamically align cross-modal features in local scales.}
AACN~\cite{xu2018attention} introduced pose-guided part attention to estimate finer part attention to exclude adjacent noise.
Without an extra pose estimator, the part-based convolutional baseline~\cite{sun2018beyond} employed a horizontal splitting strategy to extract high-level segmentation features and correct within-part inconsistency of all column vectors according to their similarities to each part.
Part-based representation learning has also proven reliable in   cross-view geo-localization~\cite{wang2021each,zhuang2021faster,dai2021transformer,chendenseLPN}.
LPN~\cite{wang2021each} uses a square-ring partition strategy to explore contextual information, which is robust to aerial image rotation.
Shen~\textit{et al.}~\cite{zhuang2021faster} designed a multi-scale partition strategy to enhance the robustness of a model to offset and scale.
FSRA~\cite{dai2021transformer} proposed a transformer-based baseline framework, and realized automatic region segmentation that increased model accuracy and robustness to position shifting.
The proposed SDPL is a part-based method inspired by   LPN~\cite{wang2021each}. The SDPL protects the global structure against scale changes while increasing the diversity of partitions to fully mine context information and  explicitly adjusts the segmentation center of the partition strategy, thus enhancing   robustness to scenarios with non-central targets.

\section{Methods}\label{Methods}
We introduce the proposed two-branch framework, including three phases: feature extraction~(section~\ref{Feature Extraction}), shifting-dense partition learning~(sections~\ref{Dense partition strategy}, \ref{Feature shifting strategy}), and classification supervision~(section~\ref{Classification supervision}).
Feature extraction acquires high-level feature maps of drone and satellite views through a backbone network.
To further extract robust feature representations for cross-view matching, we introduce the shifting-dense partition learning~(SDPL). 
SDPL includes the dense partition and shifting fusion strategies to jointly deal with target scale changes and position shifting.
After outputs of SDPL pass through the classifier, we minimize the cross-entropy loss of features from different views, so as to increase the utilization of semantic information.

\textbf{Task definition.}
Given one geo-localization dataset, we denote the input image pairs as $\{X_{D}, X_{S}\}$, where subscripts $D$ and $S$ indicate drone-  and satellite-view, respectively.
The corresponding label $Y\in[1,C]$, where $C$ indicates the   number of categories of satellite-view images.
University-1652~\cite{zheng2020university} includes 701 buildings, each containing 1 satellite-view image and 54 drone-view images.
We number 701 buildings into 701 indexes, with label $Y\in[1,701]$.
For cross-view geo-localization, the goal is to find a mapping function that projects images from different platforms to one shared semantic space, and then identify images based on feature distribution, \textit{i.e.}, images of the same location are close and vice versa.

\subsection{Feature extraction}\label{Feature Extraction}
As mentioned above, both drone- and satellite-view images are aerial-view images, with similar feature domains. We adopt one feature extractor to process the two branches, \textit{i.e.}, we use weight sharing.
SDPL can deploy various backbone networks as the feature extractor, such as ResNet~\cite{he2016deep} and SwinTransformer~\cite{liu2021swin}.
For illustration, we adopt ResNet-50~\cite{he2016deep}, pretrained on ImageNet~\cite{deng2009imagenet}, as the backbone if not specified.
We remove the average pooling and fully connected layers, and obtain intermediate feature maps for subsequent dense partition processing.
Given an input image $x_{i}\in \mathcal{R}^{3\times512\times512}$, \cq{after the backbone model we can acquire the corresponding feature map $f_{i}\in \mathcal{R}^{2048\times32\times32}$,}
\textit{i.e.},
\begin{equation}
f_i=\mathcal{F}_{backbone}\left(x_i\right), i \in\{D, S\},
\end{equation}
where $\mathcal{F}_{backbone}$ denotes the process of feature extraction, and subscripts $D$ and $S$ denote the drone-  and satellite-view, respectively.

\subsection{Dense partition strategy}\label{Dense partition strategy}
Inspired by the square-ring partition strategy~(SPS)~\cite{wang2021each}, we propose the dense partition strategy~(DPS), which recombines the basic parts of SPS to mine fine-grained features while preserving the global structure.
For a better understanding of DPS, we first briefly review SPS.

Fig.~\ref{denselpn img}(a) shows a simplified diagram of SPS, which splits the input feature into $N_{SPS}$ parts according to the distance to the image center.
In practice, SPS separates feature map $f_{i}\in \mathcal{R}^{2048\times 32\times 32}$ into $N_{SPS}$ ($N_{SPS}=4$ for illustration) part-level features
$f_{i}^{n_{SPS}}\in R^{2048\times 8n_{SPS}\times 8n_{SPS}}$~($n_{SPS}\in [1,N_{SPS}]$), where $n_{SPS}$ represents the $n$-th part.
SPS explicitly explores contextual information to improve matching accuracy, but continuous features at the boundary are inevitably divided into two parts, 
resulting in the loss of structural information.

To alleviate this limitation, we propose DPS, which gradually dilates the boundaries of partitions to allow adjacent features to be divided into the same partitions, thus mitigating local structural damage.
\cq{For instance, comparing the $P3$-th part of LPN with the $P6$-th and $P7$-th parts of DPS in Fig.~\ref{denselpn img}(a), we observe that the $P3$-th part is expanded into partitions with a wider neighborhood, meaning that more local structures are retained within the same part. 
DPS explicitly maintains global features to preserve intact structural information.
As a result, DPS can generate richer fine-grained features than SPS, thus increasing the robustness of SDPL to scale changes and target offset.}
Intuitively, DPS can be seen as an extended version of SPS, and the relationship between their numbers of partitions can be expressed as:
\begin{equation}\label{NDPS AND NSPS}
N_{DPS}=\frac{N_{SPS}(N_{SPS}+1)}{2},
\end{equation}
where $N_{SPS}$ is the number partitions of SPS and $N_{DPS}$ is the corresponding number of partitions of DPS.

Fig.~\ref{denselpn img}(a) shows a simplified diagram of DPS.
Taking partition blocks~$\{{f}_{i}^{1}$,${f}_{i}^{2}$,${f}_{i}^{3}$,${f}_{i}^{4}\}$ from SPS as the basis, DPS densely recombines the parts of the same resolution to build additional six parts~$\{f_{i}^{5}$,$f_{i}^{6}$,$f_{i}^{7}$,$f_{i}^{8}$,$f_{i}^{9}$,$f_{i}^{10}\}$.
Therefore, the outputs of DPS can be expressed as~$\{f_{i}^{1}$,...,$f_{i}^{10}\}$, with shape:
\begin{equation}
\left\{\begin{array}{l}
{f}_{i}^{1}\in \mathcal{R}^{2048 \times 8 \times 8} \\
{f}_{i}^{2},f_{i}^{5}\in \mathcal{R}^{2048 \times 16 \times 16} \\
{f}_{i}^{3},f_{i}^{6},f_{i}^{8}\in \mathcal{R}^{2048 \times 24 \times 24} \\
{f}_{i}^{4},f_{i}^{7},f_{i}^{9},f_{i}^{10}\in \mathcal{R}^{2048 \times 32 \times 32}. \\
\end{array}\right.
\end{equation}
The overall feature partition processing can be formulated as:
\begin{equation}\label{DPS process}
\Omega(f_{i}^{n_{DPS}})=\{f_{i}^{n_{DPS}} \mid \mathcal{F}_{DP}(f_{i};N_{DPS} ) \},
\end{equation}
where $\mathcal{F}_{DP}$ denotes DPS processing, ${f}_{i}^{n_{DPS}}$ denotes the $n_{DPS}$-th parts generated by DPS, and $\Omega(f_{i}^{n_{DPS}})$ is the set of all partitions.
\cq{Note that in practice, DPS recombines high-level features of various resolutions. 
For ease of understanding, we depict the recombination process as shown in Fig.~\ref{denselpn img}(b).
The low resolution part first obtains the same size as the high resolution part through a padding layer, and then sums with the latter to get the recombination feature.}
\cq{To reduce the costs of model optimization and inference, the GeM pooling layer~\cite{gu2018attention} is applied to compress each part in spatial dimensions as inputs of the subsequent classifier module,}
\begin{equation}\label{avg process}
\Omega(g_{i}^{n_{DPS}})=\{g_{i}^{n_{DPS}} \mid \mathcal{F}_{GeM}\left(\Omega(f_{i}^{n_{DPS}})\right) \}, 
\end{equation}
where $\mathcal{F}_{GeM}$ denotes GeM pooling operation, and $\Omega(g_{i}^{n_{DPS}})$ denotes the set of output features $g_{i}^{n_{DPS}} \in \mathcal{R}^{2048\times1 \times1}$.

\subsection{Shifting fusion strategy}\label{Feature shifting strategy}
The above DPS is based on the assumption that the target is located in the central regions, but this is not always the case, \textit{i.e.}, the target may deviate from the central regions.
\cq{Although the global feature can reduce the sensitivity of the part-based method to the target offset~\cite{zhuang2021faster}, we expect to further optimize DPS.}
\cq{A simple idea is to adjust the segmentation center of DPS to fit non-central target scenarios, and thus the shifting-fusion strategy is proposed}, consisting of diagonal-shifting partitioning and adaptive fusion.
The first step, diagonal-shifting partitioning, controls DPS to generate multi-group partitions based on various separation centers~(\cq{Table~\ref{DPS ABC} supports that DPS with various segmentation centers shows an-isotropic anti-offset ability}). 
\cq{Instead of preserving all parts, our adaptive fusion strategy adopts a weight estimation module to fuse partitions, thus adaptively integrating the anti-offset ability of multi-group partitions~(\textit{i.e.}, partitions generated by DPS with various segmentation points.)}

\textbf{Diagonal-shifting partitioning.}
To ensure that the segmentation center of DPS ($\mathcal{C}_{seg}^{DPS}$) does not coincide with the center of the high-level feature map $f_{i}\in \mathcal{R}^{C\times H\times W}$, we introduce the offset~($\Delta H,\Delta W$) to fine-tune $\mathcal{C}_{seg}^{DPS}$,
\begin{equation}\label{DPS center}
\mathcal{C}_{seg}^{DPS}(\Delta H,\Delta W) = (\frac{H}{2}+\Delta H,\frac{W}{2}+\Delta W),
\end{equation}
where $\{\Delta H>0 \wedge \Delta W<0\}$ and $\{\Delta H<0 \wedge \Delta W>0\}$ denote top-left and bottom-right shifting, respectively. 
Compared with unconstrained directional shifting, diagonally adjusting $\mathcal{C}_{seg}^{DPS}$ can more effectively replace the contents of the same numbered parts; hence, we set $\Delta W = -W \frac{\Delta H }{H}$.
Considering that part-level features generated by the diagonal-shifting partition strategy should have the same number of pixels to facilitate subsequent classification, we set the threshold $\frac{H}{2\cdot{N}_{SPS}}$ for the offset $\Delta H$.
Therefore, Equation~\ref{DPS center} can be optimized to:
\begin{equation}\label{segmentation point adjustment}
\mathcal{C}_{seg}^{DPS}(\Delta H) = (\frac{H}{2}+\Delta H,\frac{W}{2}-W \frac{\Delta H }{H}),
\end{equation}
where $|\Delta H| \in [0, \frac{H}{2\cdot{N}_{SPS}}]$.
Equations \ref{DPS process} and \ref{avg process} can be rewritten as:
\begin{equation}
\Omega_{\Delta H}(f_{i}^{n_{DPS}})=\{f_{i}^{n_{DPS}} \mid \mathcal{F}_{DP}(f_{i};N_{DPS};\mathcal{C}_{seg}^{DPS}(\Delta H)) \}
\end{equation}
\begin{equation}
\Omega_{\Delta H}(g_{i}^{n_{DPS}})=\{g_{i}^{n_{DPS}} \mid \mathcal{F}_{GeM}\left(\Omega_{\Delta H}(f_{i}^{n_{DPS}})\right)\}.
\end{equation}

As shown in Fig.~\ref{shifting strategy img}(a), to allow the part-level blocks to capture a target with a larger offset range, we   design three diagonal movement strategies: Centered~($\Delta H_{0}=0$), Top-Left Shifting~($\Delta H_{1} \in [0,\frac{H}{2\cdot{N}_{SPS}}]$), and Bottom-Right Shifting~($\Delta H_{2} \in [\frac{-H}{2\cdot{N}_{SPS}},0]$). 
Then the corresponding partitions~$\{\Omega_{\Delta H_{0}}(g_{i}^{n_{DPS}}), \Omega_{\Delta H_{1}}(g_{i}^{n_{DPS}}), \Omega_{\Delta H_{2}}(g_{i}^{n_{DPS}})\}$ are fed into the subsequent adaptive fusion module.
The values of $\{\Delta H_{1}, \Delta H_{2}\}$ are specified in section~\ref{Implementation Details}.

\textbf{Adaptive Fusion.}~\label{Adaptive Fusion}
To efficiently integrate the anti-offset ability of multi-group partitions, we design an adaptive fusion strategy, as shown in Fig.~\ref{shifting strategy img}(b).
We adopt a weight estimation module to learn the  probability distribution of multiple shifting partitions, $\beta_{i}\in \mathcal{R}^{1\times 3}$, from the high-level features~$f_{i}$, which can be expressed as:
\begin{equation}
\beta_{i} = \mathcal{F}_{we}(f_{i}),
\end{equation}
\cq{where $\mathcal{F}_{we}$ denotes the weight estimation module, whose CNN structure is listed in Table~\ref{weight estimation parameters}.
Assuming the input is $f_{i}$, one convolution layer is first used for the nonlinear transformation of the input, followed by an average pooling layer to obtain the feature vector.
Then we leverage two fully connected layers and a 3\textit{-ch} weight estimation layer to adequately learn the weights $\beta_{i}$ of various shifting patterns.
A softmax layer normalizes the estimated weights to optimize the network training process.}
Finally, we use $\beta_{i}$ to merge multiple sets of partitions,
\begin{equation}
\Omega\left(z_i^{n_{DPS}}\right)=\sum_{j\in {0,1,2}} \beta_i^j \cdot \Omega_{\Delta H_j}\left(g_i^{n_{DPS}}\right).
\end{equation}
To sum up, the advantages of adaptive fusion are twofold: based on the global structure information of high-level features, SDPL can adaptively highlight the optimal shifting partitions in an end-to-end manner, and merging multiple groups of partitions for subsequent classification can reduce computational costs.

\begin{table}[!t]
\centering
\caption{Structure of the weight estimation module. The parameters that need to be modified at each layer are shown in bold. 
For simplicity, we set $N=1$.}
\label{weight estimation parameters}
\resizebox{1.0\linewidth}{!}{
\begin{tabular}{lcccc}
\toprule
\multirow{2}*{Layers}  &  & Input shape & Output shape & kernel       \\ 
  &  & $N\times C\times H\times W$ & $N\times C\times H\times W$ & size     \\
\midrule
Convolution layer     &  & 1$\times$\textbf{2048}$\times$32$\times$32  & 1$\times$\textbf{1024}$\times$32$\times$32  & \textbf{1}$\times$\textbf{1}     \\
Avg pooling     &  & 1$\times$\textbf{1024}$\times$32$\times$32   & 1$\times$\textbf{1024}$\times$1$\times$1      & \textbackslash{}                \\ \hdashline
Fully connected layer &  & 1$\times$\textbf{1024}  & 1$\times$\textbf{512}    & \textbackslash{} \\
Fully connected layer &  & 1$\times$\textbf{512}   & 1$\times$\textbf{512}    & \textbackslash{} \\
Weight esitmation layer &  & 1$\times$\textbf{512}   & 1$\times$\textbf{3}      & \textbackslash{} \\ 
Softmax layer &  & 1$\times$3  & 1$\times$3   & \textbackslash{} \\ 
\bottomrule
\end{tabular}
}
\end{table}

\subsection{Classification supervision}\label{Classification supervision}
Each partition block in $\Omega\left(z_i^{n_{DPS}}\right)$ is sent to the classifier to predict the geo-tag and compare it with the ground-truth.
The classifier includes a fully connected layer, a batch normalization layer, a dropout layer, and a classification layer.
Through the classifier, the shape of each partition $z_{i}^{n_{DPS}}$ is the same as the label $Y$.
Then we apply a  cross-entropy loss function to optimize the parameters of SDPL.
The overall process can be expressed as:
\begin{equation}
O_{i}^{n_{DPS}}=\mathcal{F}_{classifier}\left(z_{i}^{n_{DPS}};C\right)
\end{equation}
\begin{equation}
\operatorname{Loss}=\sum_{i, n_{DPS}}-\log \left(\frac{\exp \left(O_{i}^{n_{DPS}}(Y)\right)}{\sum_{c=1}^C \exp \left(O_{i}^{n_{DPS}}(c)\right)}\right),
\end{equation}
where $C$ is the number of geo-tag categories. 
Note that the losses on the image of different parts and platforms are calculated separately.

\cq{\subsection{Difference from other part-based learning methods}} \label{Discussion}
\cq{In UAV-view geo-localization, several solutions based on feature segmentation have been proposed.
Similar to LPN~\cite{wang2021each}, our dense partition strategy~(DPS), one key component of SDPL, also splits feature maps by spatial dimension.
To overcome the global information loss caused by feature segmentation, DPS explicitly retains global features like MSBA~\cite{zhuang2021faster} and MBF~\cite{zhu2023uav}.
Due to the predefined shape of partitions, LPN, MSBA, and MBF have the problem of dividing continuous features into different parts.
DPS addresses this problem by subtly applying a dense combination of basic parts to generate diverse partitions, gradually expanding the boundaries of parts.
Unlike the patch-level segmentation strategy of FSRA~\cite{dai2021transformer}, the partitions of DPS contain both the target and the surrounding environment, emphasizing contextual information.
Furthermore, we discover that the part-based approaches are sensitive to the spatial changes of contextual information.
When the image content is shifted, the performance of part-based methods, such as LPN~(CNN-based) and FSAR~(Transformer-based), will decrease dramatically.
Therefore, we propose a shift fusion strategy, which further improves the anti-offset ability of center-based DPS by integrating multiple sets of partitions with various segmentation points.}

\section{Experiments and Results}\label{Experiments and Results}
\subsection{Implementation Details}\label{Implementation Details}
\textbf{Datasets}.
We verify the robustness of SPDL on two prevailing benchmarks, University-1652~\cite{zheng2020university} and SUES-200~\cite{zhu2023sues}, which are completely independent, and are described  in Table~\ref{dataset details}.

University-1652~\cite{zheng2020university} includes 1652 locations from 72 universities. 
The training set includes 701 buildings of 33 universities, each with 1 satellite-view, 54 drone-view images, and several street-view images, where the latter are not applied in this paper.
The testing set includes the other 951 buildings of the remaining 39 universities. 
There are no overlapping universities in the training and test sets.
The testing set is further divided according to the requirements of the sub-tasks,~\textit{i.e.}, drone-view target localization (Drone$\rightarrow$Satellite) and drone navigation (Satellite$\rightarrow$Drone).
For the drone-view target localization sub-task, there are 37854 drone-view images~(as Query), 701 true-matched satellite-view images~(as Gallery), and 250 satellite-view distractors.

  SUES-200~\cite{zhu2023sues} is a cross-view matching dataset with only two views, drone-view and satellite-view. 
Different from University-1652, SUES-200 focuses on the impact of scene range on cross-view geo-location by collecting drone-view images at heights of 150, 200, 250, and 300 m. 
SUES-200 includes 200 locations, each with 50 drone-view images and 1 corresponding satellite-view image at a fixed altitude~(there are  200$\times$50$\times$4 drone images, and 200 satellite images).
SUES-200 is divided into training and test sets, with 120 locations for training, and the remaining 80 locations as test data.
To increase the difficulty of matching, the gallery set of each sub-task is composed of testing and training data, the latter for confusion.

\begin{table}[!t]
\centering
\caption{Statistics of two datasets, including the image number of query set and gallery set.}
\label{dataset details}
\resizebox{1.0\linewidth}{!}{
\begin{tabular}{lccccccccc}
\toprule
\multirow{3}{*}{Dataset} &  & \multicolumn{2}{c}{Training phase}                    &  & \multicolumn{5}{c}{Test phase}                                     \\ \cline{3-4} \cline{6-10} 
                         &  & \multicolumn{2}{c}{\multirow{2}{*}{Drone $\leftrightarrow$ Satellite}} &  & \multicolumn{2}{c}{Drone$\rightarrow$Satellite} &  & \multicolumn{2}{c}{Satellite$\rightarrow$Drone}       \\
                         &  & \multicolumn{2}{c}{}                                  &  & Query      & Gallery      &  & Query            & Gallery          \\
                         \midrule
University-1652~\cite{zheng2020university}          &  & 37854                     & 701                       &  & 37854      & 951          &  & 701              & 51354            \\ \hline
SUES-200~\cite{zhu2023sues}                 &  & 24000                     & 120                       &  & 16000      & 200          &  & 80               & 40000            \\ 
\bottomrule
\end{tabular}
}
\end{table}

\begin{table}[!t]
\centering
\caption{The classifier module parameters under different datasets and different backbones. $<$$\cdot$, $\cdot$$>$ denotes the input and output channels of each learnable layer.}
\label{model paremeters}
\resizebox{1.0\linewidth}{!}{
\begin{tabular}{lcccccc}
\toprule
\multirow{3}{*}{Datasets} &  & \multicolumn{5}{c}{Backbones}                                  \\ \cline{3-7} 
                          &  & \multicolumn{2}{c}{ResNet-50~\cite{he2016deep}} && \multicolumn{2}{c}{SwinV2-B~\cite{he2016deep}} \\ \cline{3-4} \cline{6-7} 
                          && $F_C$      & $F_{Cls}$          && $F_C$       & $F_{Cls}$       \\ 
                          \midrule
University-1652~\cite{zheng2020university} &&   $<$2048, 512$>$     &   $<$512, 701$>$    &&  $<$1024, 512$>$   &   $<$512, 701$>$    \\ \hline
SEU-200~\cite{zhu2023sues} &&   $<$2048, 512$>$    &    $<$512, 120$>$           &&  $<$1024, 512$>$    &   $<$512, 120$>$       \\ 
\bottomrule
\end{tabular}
}
\end{table}

\textbf{Model Details}.
For the proposed SDPL, we set the number of  partitions  to $N_{DPS}$=10, with  offset values $\Delta H_{1}$=2 and $\Delta H_{2}$=-2.
In ablation experiments, we uniformly employ   ResNet-50~\cite{he2016deep} with pretrained weights on ImageNet~\cite{krizhevsky2012imagenet} to extract visual features, and   modify the stride of the second convolutional layer and the last down-sample layer in conv5\_1 of the ResNet-50 from 2 to 1.
In the comparison experiment, SwinV2-B~\cite{liu2022swin} is also adopted as the backbone to verify the compatibility between SDPL and the transformer structure.
The classifier module contains two linear layers~(a feature compression layer $F_C$, and a feature classification layer $F_{Cls}$) whose parameters are shown in Table~\ref{model paremeters}.
The input channel of   $F_C$ is consistent with the output feature of the backbone, \textit{i.e.}, 2048 and 1024 for ResNet-50 and SwinV2-B, respectively, while the output channel of   $F_{Cls}$ is related to  the training set of University-1652 and SUES-200, \textit{i.e.}, 701 and 120.
During the test, we extract the part-level image representation before the classification layer $F_{Cls}$.
Then we concatenate part-level features as the final visual descriptor of the input image, and the Euclidean distance  to measure the similarity between the query   and candidate images in the gallery.

\begin{table*}[!t]
\centering
\caption{\cq{Comparison with the state-of-the-art results reported on University-1652. The compared method are divided into three groups, the Resnet50-based methods at the top, the ViT-based methods at the middle and the methods based on other backbones~(\textit{e.g.}, Swin and OSNet) are at the bottom.
Best and second best performance are in red and blue colors.}}
\label{University1652 Results}
\resizebox{0.95\linewidth}{!}{
\begin{tabular}{ccccccccccccc}
\hline
 &  &            &     &        &      &        &        & \multicolumn{5}{c}{University-1652}   \\ \cline{9-13} 
 &  &         &         &      &        &       &         & \multicolumn{2}{c}{Drone-Satellite}    &       & \multicolumn{2}{c}{Satellite-Drone}   \\ \cline{9-10} \cline{12-13} 
\multirow{-3}{*}{Method}        &  & \multirow{-3}{*}{Publication}    &      & \multirow{-3}{*}{Backbone}    &     & \multirow{-3}{*}{Image size} &    & Recall@1    & AP      &   & Recall@1     & AP    
\\ \hline 
Instance Loss~\cite{zheng2020university}       &  & ACMMM' 2020      &    & ResNet-50    &       & 512$\times$512     &   & 59.69  & 64.80    &   & 73.18    & 59.40      
\\
LCM~\cite{ding2020practical}   &  & Remote Sens' 2020     &     & ResNet-50      &   & 512$\times$512      &          & 66.65       & 70.82         &            & 79.89    & 65.38   
\\ 
PCL~\cite{tian2021uav}       &  & TCSVT' 2021       &   & ResNet-50     &  & 512$\times$512        &       & \textcolor{blue}{83.27}    & \textcolor{blue}{87.32}       &    & \textcolor{red}{91.78}    & \textcolor{blue}{82.18}         
\\
LPN~\cite{wang2021each}        &  & TCSVT' 2021          &          & ResNet-50       &        & 512$\times$512       &        & 77.71     & 80.80       &       & 90.30         & 78.78        
\\
RK-Net~\cite{lin2022joint}    &  & TIP' 2022    &        & ResNet-50          &      & 512$\times$512        &           & 68.10      & 71.53      &        & 80.96  & 69.35   
\\
CA-HRS~\cite{lu2022content}    &  & ACCV' 2022     &  & ResNet-50   &       & 512$\times$512     &            & 81.00     & 83.64        &     & 88.16    & 77.46 
\\
MCCG~\cite{shen2023mccg}     &  & TCSVT' 2023    &        & ResNet-50     &                          & 512$\times$512        &    & 70.94      & 75.04      &                          & 83.02     & 69.36     \\
Sample4Geo~\cite{deuser2023sample4geo}      &  & ICCV' 2023    &    & ResNet-50      &       & 512$\times$512     &       & 78.62    & 82.11     && 87.45    & 76.32     
\\
AEN(w. LPN)~\cite{liu2024adaptive}     &  & SPL' 2024 &  & ResNet-50 &  & 256$\times$256  &  & 77.40 & 80.27 &  & \textcolor{blue}{90.30} & 76.01
\\
SDPL    && --   &      & ResNet-50  && 512$\times$512  &&  \textcolor{red}{85.19}  &  \textcolor{red}{87.43}     &&   89.30   &  \textcolor{red}{82.75} 
\\
\hdashline
ViT~\cite{dosovitskiy2020image}    && ICLR' 2020    &  & ViT-S &    & 512$\times$512   &&  74.09  &   77.82  &&   83.31  &   72.27    \\
FSRA~\cite{dai2021transformer}    &  & TCSVT' 2021      &        & ViT-S      &  & 512$\times$512     &       & 85.50      & 87.53     && \textcolor{blue}{89.73}     & \textcolor{blue}{84.94}    \\
Dai~\cite{dai2023vision}       &  & TIP' 2023       &         & ViT-S      &     & 224$\times$224       &        & 82.22        & 84.78      &      & 87.59   & 81.49      \\
TransFG~\cite{zhao2024transfg} &  & TGRS' 2024    &   & ViT-S   &      & 512$\times$512   &     & \textcolor{red}{87.92}   & \textcolor{red}{89.99}    &  & \textcolor{red}{93.37}   & \textcolor{red}{87.94} \\ 
SDPL     && --      && ViT-S     && 512$\times$512  &&  \textcolor{blue}{85.57}       &   \textcolor{blue}{87.61}   &&   88.73  &  84.75  \\ 
\hdashline
Swin-B~\cite{liu2021swin}   && CVPR' 2021   && Swin-B   &   & 256$\times$256     && 84.15    & 86.62    && 90.30    & 83.55        \\
SwinV2-B~\cite{liu2022swin}        && CVPR' 2022    && SwinV2-B  &   & 256$\times$256       &     & 86.99     & 89.02    &      & 91.16      & 85.77        \\ 
F3-Net~\cite{sun2023f3}     && TGRS' 2023     && --   &        & 384$\times$384    && 78.64     & 81.60        &         & --          & --     \\
Song~\cite{10288351}     && GRSL' 2023     &&  OSNet   &        & 512$\times$512    &&   83.26   &    85.84     &&   90.30        &  82.71   \\
MFJRLN(Lcro)~\cite{ge2024multi}        &  & TGRS' 2024 &     & Swin-B   &          & 224$\times$224   &    & 87.61      & 89.57    &      & 91.07       & 86.72  \\
GeoFormer~\cite{li2024geoformer}          &  & J-STARS' 2024    &     & E-Swin-B   &    & 224$\times$224   &     & \textcolor{blue}{88.16}    & \textcolor{blue}{90.03}    &  & \textcolor{blue}{91.87}    & \textcolor{blue}{87.92}       \\
SDPL     &  & --     &     & SwinV2-B      &  & 256$\times$256      &       & \textcolor{red}{90.16}   &  \textcolor{red}{91.64}     &&   \textcolor{red}{93.58}   &   \textcolor{red}{89.45}      \\ \hline
\end{tabular}
}
\end{table*}

\begin{table*}[!t]
\centering
\caption{\cq{Comparison with the state-of-the-art results reported on SEUS-200~\cite{zhu2023sues} dataset. The input image size is $256\times256$.}}
\label{SEUS200 Results}
\resizebox{0.90\linewidth}{!}{
\begin{tabular}{ccccccccccccc}
\toprule
\multicolumn{13}{c}{Drone$\rightarrow$Satellite}                          \\ \hline
\multirow{2}{*}{Method} &&  \multicolumn{2}{c}{150m} &&  \multicolumn{2}{c}{200m} &&  \multicolumn{2}{c}{250m} && \multicolumn{2}{c}{300m} \\
                        && R@1    & AP    && R@1     & AP      && R@1    & AP     && R@1      & AP         \\ 
                        \midrule
SUES-200~\cite{zhu2023sues}   &&  55.65    &   61.92  &&  66.78  &  71.55  && 72.00   &   76.43  &&   74.05  &  78.26    \\
LCM~\cite{ding2020practical}  &&  43.42   &   49.65   && 49.42   & 55.91  &&  54.47    & 60.31   &&  60.43    &   65.78     \\
LPN~\cite{wang2021each}  &&  61.58  &  67.23  && 70.85  & 75.96   &&    80.38   &  83.80      &&  81.47    &  84.53   \\
FSRA~\cite{dai2021transformer}      &&   68.25   &   73.45  &&  83.00  &   85.99  && 90.68          &  92.27   &&   91.95   &   93.46    \\
MBF~\cite{zhu2023uav}   &&   \textcolor{red}{85.62}   &   \textcolor{red}{88.21}  &&   87.43  &   90.02  && 90.65   & 92.53   &&   92.12   &   93.63    \\
MCCG~\cite{shen2023mccg}    &&   82.22   &   85.47  &&  {89.38}   &   \textcolor{blue}{91.41}   &&  \textcolor{blue}{93.82}     &  \textcolor{blue}{95.04}   &&   {95.07}    &    {96.20}  \\ 
MJRLIFS~\cite{ge2024multibranch}    &&   77.57   &   81.30  && \textcolor{blue}{89.50}   &   91.40   && 92.58     &  94.21   && \textcolor{blue}{97.40}    &    \textcolor{blue}{97.92}  \\ 
\textbf{SDPL} &&    \textcolor{blue}{82.95}  &  \textcolor{blue}{85.82}      && \textcolor{red}{92.73}   &  \textcolor{red}{94.07}  &&    \textcolor{red}{96.05}   &   \textcolor{red}{96.69}  &&   \textcolor{red}{97.83}   &  \textcolor{red}{98.05}     \\
\bottomrule
\multicolumn{13}{c}{Satellite$\rightarrow$Drone}                          \\ \hline
\multirow{2}{*}{Method}  && \multicolumn{2}{c}{150m} && \multicolumn{2}{c}{200m} && \multicolumn{2}{c}{250m} && \multicolumn{2}{c}{300m} \\
                        && R@1  & AP  && R@1  & AP   && R@1    & AP  && R@1  & AP     \\ \midrule
SUES-200~\cite{zhu2023sues}  && 75.00  &  55.46  &&  85.00  & 66.05  &&  86.25   &   69.94  &&  88.75  &  74.46   \\
LCM~\cite{ding2020practical}  &&  57.50   &  38.11  &&  68.75    &   49.19   && 72.50   & 47.94  &&  75.00   &  59.36      \\
LPN~\cite{wang2021each}  &&  83.75  &  66.78  &&  88.75 & 75.01           &&   92.50   &   81.34  &&   92.50    &   85.72   \\
FSRA~\cite{dai2021transformer}  &&  83.75  &   76.67  &&  90.00   &  85.34  &&  93.75 &  90.17    &&  95.00   &   92.03  \\
MBF~\cite{zhu2023uav}   &&   88.75   &   \textcolor{blue}{84.74}  &&  91.25  &  89.95  &&  93.75   & 90.65   &&  96.25   &  91.60   \\
MCCG~\cite{shen2023mccg}    &&   {93.75}   &   \textcolor{red}{89.72}  &&   \textcolor{blue}{93.75}    &   \textcolor{blue}{92.21}  &&   {96.25}   &  \textcolor{red}{96.14}   &&    \textcolor{blue}{98.75}   &   \textcolor{blue}{96.64}  \\ 
MJRLIFS~\cite{ge2024multibranch}   &&  \textcolor{blue}{93.75}  &   79.49  &&  \textcolor{red}{97.50} &  90.52  &&  \textcolor{blue}{97.50}   &  \textcolor{blue}{96.03}   &&  \textcolor{red}{100.00}   & \textcolor{red}{97.66}   \\
\textbf{SDPL}    &&   \textcolor{red}{93.75}   &   83.75   &&   \textcolor{blue}{96.25}     &   \textcolor{red}{92.42}    &&  \textcolor{red}{97.50}  &    {95.65}   &&   {96.25} &   {96.17}  \\ 
\bottomrule
\end{tabular}
}
\end{table*}


\textbf{Training Details}.
By default, we resize the input images to 512$\times$512 for both training and testing phases.
In training, we employ random horizontal image flipping as data augmentation. 
Both models are trained using stochastic gradient descent~(SGD), with momentum 0.9 and weight decay 0.0005, for 1000 epochs, with a mini-batch of 4.
The learning rate is initialized as $1\times10^{-3}$, and decayed by 0.75 after 50 epochs.
Following previous works~\cite{zheng2020university,wang2021each}, we utilize the Euclidean distance to measure the similarity between the query and candidate images in the gallery during testing phase.
Experiments are conducted using PyTorch~\cite{paszke2019pytorch} on an NVIDIA RTX 3090 GPU with 24 GB of memory.

\textbf{Evaluation Metrics}.
We adopt the Recall@K and average precision~(\textbf{AP}) to evaluate our model, which are common evaluation metrics in the cross-view geo-localization task. 
Recall@K refers to the ratio of the number of matched images in the top-K ranking list, and a higher recall score shows  better network performance.
AP measures the precision of the retrieval system.
The area under the Precision-Recall curve, \textit{i.e.}, the average precision (\textbf{AP}), is also calculated to reflect the precision and recall rate of retrieval performance.

\subsection{Comparison with State-of-the-art}\label{Comparison with the State-of-the-arts}
\textbf{Results on University-1652.}
As shown in Table~\ref{University1652 Results}, we compare the proposed SDPL with other methods on University-1652~\cite{zheng2020university}.
\cq{The quantitative results are divided into three groups, ResNet-based, ViT-based, and other backbone-based methods.}
For ResNet-based methods, SPDL achieves 85.19\% Recall@1 accuracy, 87.43\% AP on the drone-view target localization task~(Drone$\rightarrow$Satellite), 89.30\% Recall@1 accuracy, and 82.75\% AP on the drone navigation task~(Satellite$\rightarrow$Drone).
The performance has surpassed the reported results of other   methods, such asLCM~\cite{ding2020practical}, MCCG~\cite{shen2023mccg}, and RK-Net~\cite{lin2022joint}.
In particular, the Recall@1 of our method are nearly 7.48\% higher than that of LPN~\cite{wang2021each}, the most similar comparison method.
\cq{On the ViT-based track, SPDL is compatible with the ViT framework~\cite{dosovitskiy2020image}, and it achieves competitive results with FSRA~\cite{dai2021transformer}.
SDPL achieves competitive performance to TransFG~\cite{zhao2024transfg}, which is specific to ViT.
On the other backbone-based~(\textit{e.g.}, Swin~\cite{liu2021swin}) track, SDPL is also superior to most comparison methods.
Specifically, compared with the recent GeoFormer~\cite{li2024geoformer}, the proposed method improves Recall@1 and AP by 2.00\% and 1.61\%, respectively,  on Drone$\rightarrow$Satellite, and by 1.71\% and 1.53\%   on Satellite$\rightarrow$Drone.}
When injecting SDPL into SwinV2-B~\cite{liu2022swin}, AP accuracy for both tasks, \textit{i.e.}, drone-view target localization and drone navigation, improves dramatically, from 89.02\% to 91.64~(+2.62\%) and from 85.77 to 89.45~(+3.68\%), respectively. 
This demonstrates that the proposed SDPL is compatible with the transformer architecture.

\begin{figure*}[!t]
    \centering
    \includegraphics[width=1.0\linewidth]{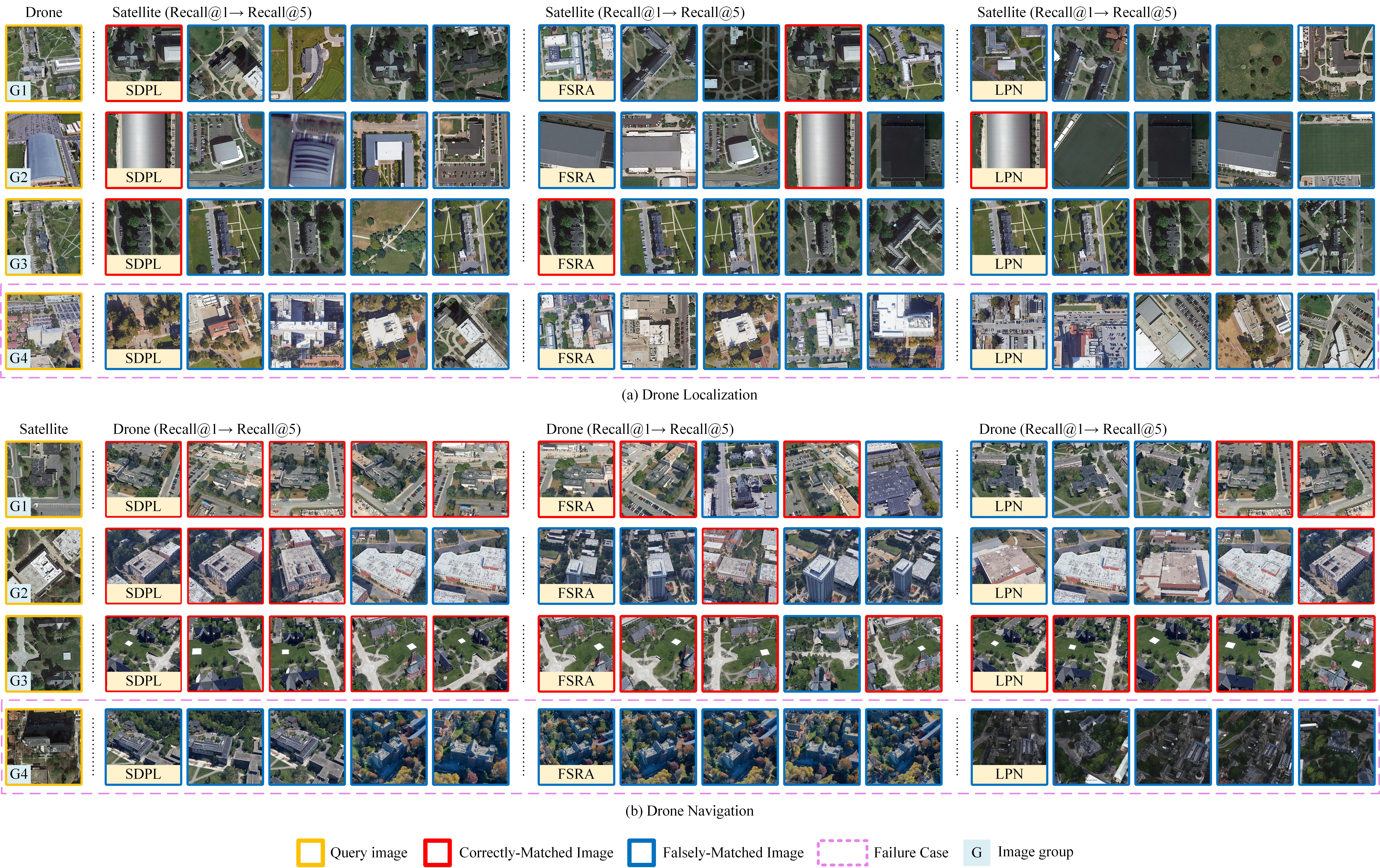}
    \caption{Image retrieval results obtained with SDPL, FSRA and LPN. (a) Top-5 retrieval results of drone localization on University-1652; (b) Top-5 retrieval results of drone navigation on University-1652.}
    \label{DSSD visual_results}
\end{figure*}

\begin{figure}[!t]
    \centering
    \includegraphics[width=0.85\linewidth]{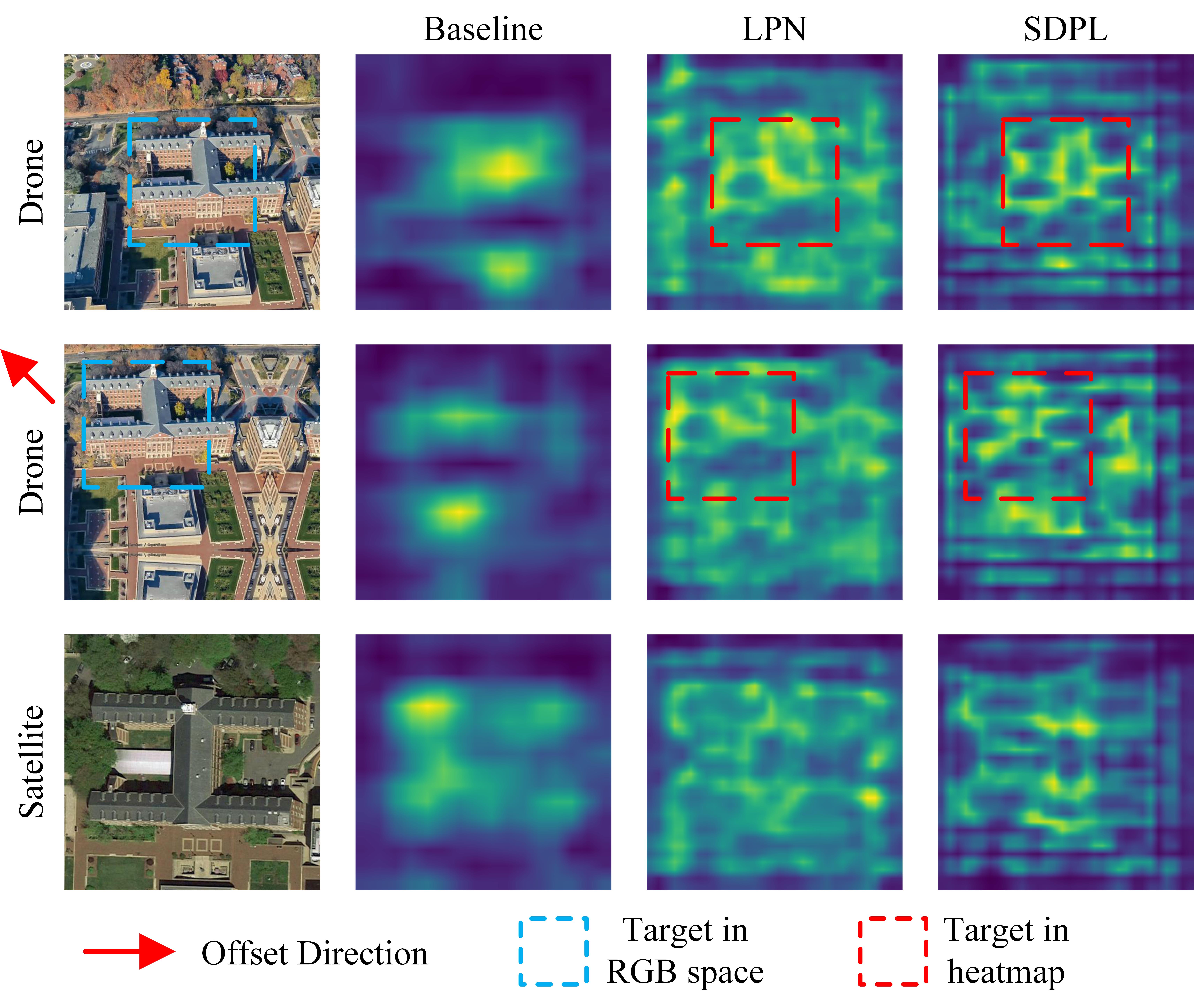}
    \caption{Visualization of heatmaps generated by Baseline, LPN and SDPL.}
    \label{heatmap_visual}
\end{figure}

\begin{table}[!t]
\centering
\caption{Performance comparison of SDPL framework with different settings. The SPS and DPS adopt the default central point for partitioning.}
\label{model structure}
\resizebox{1.0\linewidth}{!}{
\begin{tabular}{lcccccccccccc}
\toprule
\multirow{4}{*}{Model} & \multicolumn{6}{c}{Model structure}  &  & \multicolumn{5}{c}{Results}      \\ \cline{2-7} \cline{9-13} 
& \multicolumn{3}{c}{Partition strategy}     &  & \multicolumn{2}{c}{Fusion strategy} &  & \multicolumn{2}{c}{Drone$\rightarrow$Satellite}   &  & \multicolumn{2}{c}{Satellite$\rightarrow$Drone}  \\ \cline{2-4} \cline{6-7} \cline{9-10} \cline{12-13} 
& \multirow{2}{*}{None} & SPS & DPS &  &  Hard       & Adaptive       && \multirow{2}{*}{Recall@1} & \multirow{2}{*}{AP} && \multirow{2}{*}{Recall@1} & \multirow{2}{*}{AP} \\
&      &   (center)  &  (center)   &&  fusion   & fusion   &&     &     &&          &        \\ \midrule
Model-1    & $\checkmark$   &       &       &&   
&        &&       57.62    &   42.61    &&    68.05   &   55.00     \\
Model-2  &    & $\checkmark$   &         &&                     &            &&     81.59      &    84.18    &&   \textcolor{blue}{91.01}     &   81.06     
\\
Model-3  &   &    & $\checkmark$     &&      &     &&  \textcolor{blue}{85.08}     &  \textcolor{blue}{87.31}          &&     \textcolor{red}{90.73}          &   \textcolor{red}{83.43}     
\\
\hdashline
Model-4    &    &  & $\checkmark$    &&     $\checkmark$  &    &&       81.06     &   83.82  &&    86.59  &    78.30      
\\
SDPL    &     &    & $\checkmark$  &&      &      $\checkmark$    &&   \textcolor{red}{85.19}           &    \textcolor{red}{87.43}          &&   89.30     &   \textcolor{blue}{82.75}        
\\
\bottomrule
\end{tabular}
}
\end{table}

\textbf{Results on SUES-200.}
Table~\ref{SEUS200 Results} compares results with those of recent competitive methods on SUES-200~\cite{zhu2023sues}.
Notably,  all results of those methods are independent of University-1652~\cite{zheng2020university}.
Obviously, SDPL surpasses other methods on various indicators, with  excellent and stable performance at different flight heights.
In Drone$\rightarrow$Satellite, compared with MCCG~\cite{shen2023mccg} at four heights, our method boosts the Recall@1 from 82.22\%, 89.38\%, 93.82\%, 95.07\% to 82.95\%~(+0.73\%), 92.73\%~(+3.35\%), 96.05\%~(+2.23\%), 97.83\%~(+2.76\%), respectively.
In Satellite$\rightarrow$Drone, compared with MCCG~\cite{shen2023mccg} at heights of 200m and 250m, SDPL improves Recall@1 from 93.75\%, 96.25\% to 96.25\%~(+2.5\%), 97.50\%~(+1.25\%), respectively.

\textbf{Qualitative Results.}
\cq{As an additional qualitative evaluation, we visualize some retrieval results created by our and compared methods.
Fig.~\ref{DSSD visual_results} shows image retrieval results generated
by SDPL, FSRA~\cite{dai2021transformer}, and LPN~\cite{wang2021each} for drone-view localization~(Drone$\rightarrow$Satellite) and drone navigation~(Satellite$\rightarrow$Drone).
When the object in the query image is clear and prominent, SPDL can learn the distinctive features while mining the contextual information, thus retrieving reasonable images based on the content~(see Groups~1 and~2 in Figs.~\ref{DSSD visual_results}(a) and (b)).
Compared to LPN, our dense partition strategy increases the diversity of partitions~(\textit{i.e.}, more fine-grained features), while preserving global features, so SDPL has a stronger ability to distinguish similar content.
As shown in Group~3 of Fig.~\ref{DSSD visual_results}, our solution can weaken the interference of similar backgrounds, resulting in superior retrieval results.
Our failure case is shown in Group~4, in which all methods cannot find the matched image.
We summarize two reasons: a complicated background and unapparent target will dramatically increase the difficulty of image matching; and the sharpness of query images, as well as the color similarity of the cross-view, can affect the accuracy of the algorithm.}

\cq{Moreover, we visualize some heatmaps generated by Baseline~\cite{zheng2020university}, LPN~\cite{wang2021each}, and SDPL in the drone and satellite platforms, as shown in Fig.~\ref{heatmap_visual}.
We observe that the distribution of the active region is similar to that of the target in the RGB space.
Compared with the baseline, both LPN and SDPL employ partition strategies to mine fine-grained features, thus activating the region of the geographic target and neighbor areas containing contextual information.
We note that the regions activated by LPN and SDPL reflect some structural features, rather than solely focusing on target positions like the baseline.
In addition, benefiting from more fine-grained features created by DPS, the heatmaps generated by SPDL show clearer edge structures, indicating a more complete understanding of the query image.
We argue that fine-grained features are a crucial factor for the improvement of image retrieval performance.
An interesting phenomenon is that when we offset the query image, the activation region will shift along a similar direction~(red and blue boxes, Fig.~\ref{heatmap_visual}), which inspires us to adjust the segmentation point to improve the  anti-offset capability of DPS.
Compared with LPN of a fixed segmentation point, SPDL is a fusion of multiple groups of partitions, which is more robust for targeting non-central scenarios~(see following section for experimental proof).}

\subsection{Ablation Studies}\label{Ablation Studies}
To verify the effectiveness of SDPL, we conduct extensive ablation studies, including: 
\begin{itemize}
    \item To prove the effectiveness of the proposed dense partition strategy and feature shifting strategy;
    \item \cq{To prove that SDPL is robust to target non-centered scenarios, and reveal the source of anti-offset property;}
    \item \cq{To explore how SDPL leverages contextual information to improve retrieval performance~(\textit{i.e.}, to explore the effect of various combinations of fine-grained features on performance);}
    \item To select optimal parameter settings, \textit{i.e.}, offset values, number of partitions, and input image size.
\end{itemize}
Note that experiments are performed on University-1652~\cite{zheng2020university}.

\cq{\textbf{Effectiveness of each component.}
The core components of SPDL are DPS and SFS.}
To verify the effectiveness of DPS for cross-view tasks, we constructed three derived models for ablation experiments.
Model-1 has no partition strategy, \textit{i.e.}, the backbone output is fed into the classification module through a global pooling layer.
Model-2 is constructed with a square-ring partition strategy~\cite{wang2021each}, and Model-3 is constructed with DPS.
Note that Models-{1,2,3} all adopt ResNet-50~\cite{he2016deep} as a backbone, without a feature shifting strategy.
Referring to Table~\ref{model structure}, removing partition strategy~(Model-1) substantially reduces the Recall@1 and AP performances in two sub-tasks.
Comparing SPS and DPS, the latter produces more finer-grained parts to fully mine the spatial feature information, and thus achieves better results.

As mentioned above, we design a shifting-fusion strategy to mitigate performance degradation caused by non-centered target regions.
To investigate the effectiveness of the adaptive fusion strategy, we  construct the Model-4, which consists of DPS and a hard fusion strategy~(\textit{i.e.}, to manually fix the fusion ratio of the multiple groups of DPS). 
In this experiment, we set the proportions of Centered, Top-Left, and Bottom-Right shifting partitions to 0.8, 0.1, and 0.1, respectively.
Fig.~\ref{shifting strategy img} shows an illustration of the various fusion strategies applied to DPS.
Compared with the hard fusion strategy, our adaptive fusion strategy learns the proportion of multiple sets of DPS through a weight estimation module, which is more flexible, and reduces the labor cost of parameter adjustment.
Referring to Table~\ref{model structure}, the hard-shifting strategy obviously damages the accuracy on the UAV-view localization sub-task.
When replacing the former with our adaptive fusion strategy, SDPL can improve Recall@1 accuracy from 81.06\% to 85.19\%~(+4.13\%).
\cq{For the default scenarios, SDPL exhibits similar performance to Model-3. We emphasize that the purpose of SFS is to enhance the model's ability to handle position deviations, so the quantitative results of SDPL in the default scenarios are acceptable.
Subsequent experiments under non-central scenarios will highlight the benefits of our SFS.}

\begin{figure}[!t]
    \centering
    \includegraphics[width=1.0\linewidth]{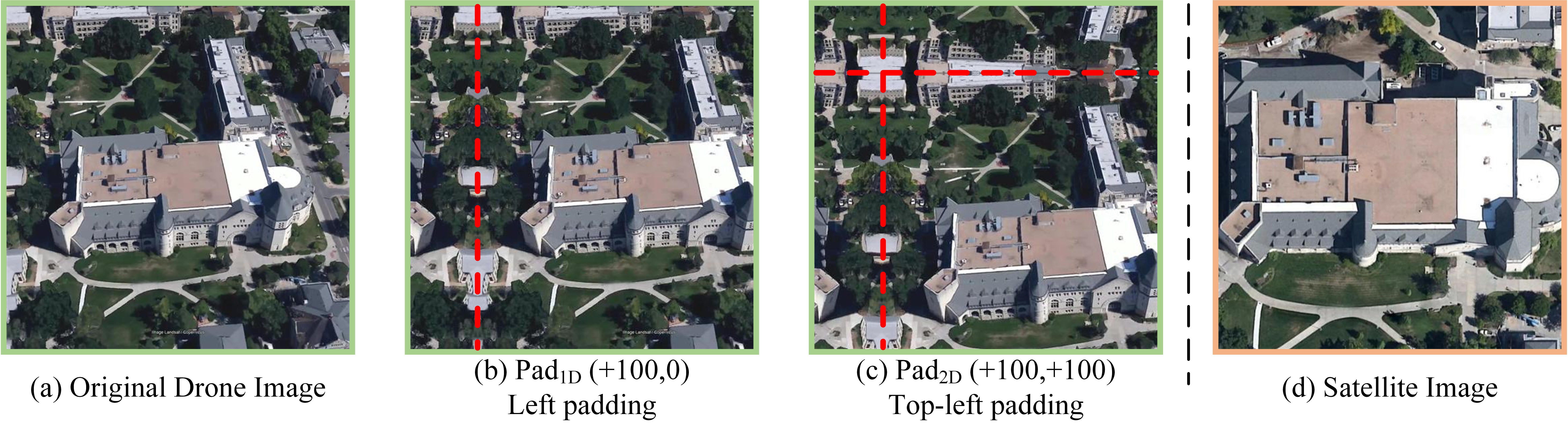}
    \caption{$Pad_{1D}$ and $Pad_{2D}$.
    (a) Original drone image; 
    (b) Image   obtained by mirroring and expanding  100-pixel-wide portion of   left side of   image and cutting off   equal pixel width on   right side;
    (c) Strips   100-pixels wide   mirrored on   top and left sides of   original image, cropped to   initial resolution;
    (d) Corresponding satellite image.
    Left space of   red dotted line consists of   extra padded pixels.}
    \label{padding pattern}
\end{figure}

\begin{figure*}[!t]
    \centering
    \includegraphics[width=1.0\linewidth]{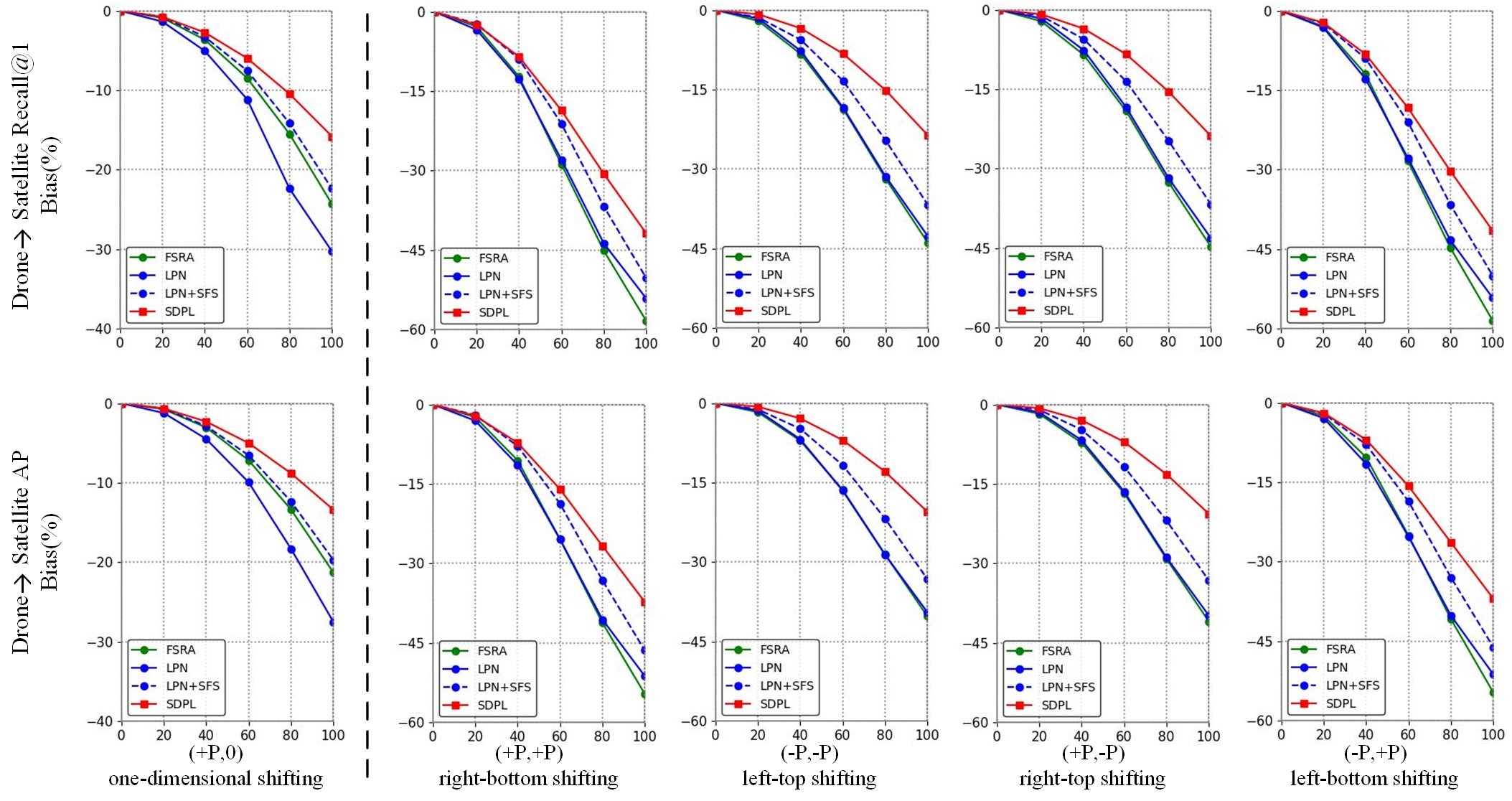}
    \caption{\cq{Impact of   number of   1-D   and   2-D pads on AP and Recall@1. Vertical axis shows   magnitude of   decrease in accuracy.}}
    \label{padding result}
\end{figure*}

\cq{\textbf{Robustness of SDPL to non-central scenarios.}}
To fully verify the robustness of the proposed SDPL for non-centered target scenarios, we manually offset query images in two shifting patterns: $Pad_{1D}$ and $Pad_{2D}$.
As shown in Fig.~\ref{padding pattern}, $Pad_{1D}$ mirrors the part with width $P$ on the left (or right) side of the image, and crops the padded image, and $Pad_{2D}$ mirrors the parts with width $P$ along the height and width dimensions of the image, and crops the padded image.
\cq{We compare the proposed SDPL with FSRA~\cite{dai2021transformer} and LPN~\cite{wang2021each}, with   results as shown in Table~\ref{padding results LPN and SDPL}.
The performance degradation under different padding patterns is shown in Fig.~\ref{padding result}.}
\cq{Obviously, AP and Recall@1 of SPDL decrease much more slowly than those of LPN and FSRA as the padding pixels increase, regardless of $Pad_{1D}$ and $Pad_{2D}$.}
While the transformer architecture can capture global dependencies, the performance of FSRA~(ViT-based) degrades significantly compared to SPDL when the target is shifted.
In particular, when $P=100$, the AP values of FSRA, LPN and SDPL are reduced by
21.29\%, 27.49\%, and 13.38\%, respectively, in the $Pad_{1D}$ pattern~($+P$,0), and by 54.61\%, 51.24\%, and 37.27\% in the $Pad_{2D}$ pattern~($+P$,$+P$).
Meanwhile, the diagonal-shifting partition strategy in SDPL only includes the top-left and bottom-right patterns, but it also has good generalization ability for the top-right~($-P$,$+P$) and bottom-left~($+P$,$-P$) shifted patterns.
Note that under the same $P$, since $Pad_{2D}$ produces larger flip regions than $Pad_{1D}$, it is logical that the former causes more decay.

\begin{table*}[!t]
\centering
\caption{\cq{Ablation study on shifting query images during inference on University-1652. We report the retrieval results and performance degradation for FSRA, LPN, LPN+SFS and SPDL in five padding patterns. Best and second best performance are in red and blue colors, respectively.}}
\label{padding results LPN and SDPL}
\resizebox{1.0\linewidth}{!}{
\begin{tabular}{lcccccccccccc}
\hline
Padding     &  & \multicolumn{2}{c}{FSRA(ViT)} &  & \multicolumn{2}{c}{LPN(ResNet)} &  & \multicolumn{2}{c}{LPN+SFS(ResNet)} &  &  \multicolumn{2}{c}{SDPL(ResNet)} 
\\ \cline{3-4} \cline{6-7} \cline{9-10} \cline{12-13} 
Pixel(P)    &  & Recall@1      & AP       &  & Recall@1     & AP           &  & Recall@1    & AP       &  & Recall@1     & AP     
\\ \hline
(0,0)       &  & \textcolor{red}{86.41$_{-0}$}      & \textcolor{red}{88.34$_{-0}$}    &  & 77.99$_{-0}$        & 80.84$_{-0}$        &  & {82.12$_{-0}$}               & {84.64$_{-0}$}     &  &  \textcolor{blue}{85.19$_{-0}$}      & \textcolor{blue}{87.43$_{-0}$}      
\\
\hdashline
(+20,0)     &  &    \textcolor{red}{85.51$_{-0.90}$}    &    \textcolor{red}{87.59$_{-0.75}$}     &  &   76.64$_{-1.36}$  &   79.62$_{-1.22}$     &  &      81.37$_{-0.75}$   &   83.98$_{-0.66}$     &  &      \textcolor{blue}{84.42$_{-0.77}$}   &    \textcolor{blue}{86.78$_{-0.65}$} 
\\
(+40,0)     &  &     \textcolor{red}{82.77$_{-3.64}$}  &   \textcolor{red}{85.30$_{-3.04}$}   &  &   72.94$_{-5.05}$  &  76.36$_{-4.48}$  &  &    78.84$_{-3.28}$    &  81.78$_{-2.86}$     &  &     \textcolor{blue}{82.46$_{-2.73}$}  &   \textcolor{blue}{85.13$_{-2.3}$}   
\\
(+60,0)     &  &  \textcolor{blue}{77.95$_{-8.46}$}  &    \textcolor{blue}{81.18$_{-7.16}$}    &  &    66.85$_{-11.14}$   &  70.91$_{-9.93}$   &  &   74.59$_{-7.53}$   &     78.10$_{-6.54}$     &  &  \textcolor{red}{79.22$_{-5.97}$} &  \textcolor{red}{82.43$_{-5.00}$}  
\\
(+80,0)     &  &  \textcolor{blue}{70.90$_{-15.51}$}   &   \textcolor{blue}{74.99$_{-13.35}$}   &  &   55.61$_{-22.38}$  &   62.55$_{-18.29}$   &  &    67.96$_{-14.16}$  &      72.29$_{-12.35}$   &  &    \textcolor{red}{74.73$_{-10.46}$}  &   \textcolor{red}{78.63$_{-8.80}$}   
\\
(+100,0)    &  &  \textcolor{blue}{62.10$_{-24.31}$}   &  \textcolor{blue}{67.05$_{-21.29}$}   &  &   47.73$_{-30.26}$    &  53.35$_{-27.49}$   &  &  59.73$_{-22.39}$    &     64.93$_{-19.71}$  &  &     \textcolor{red}{69.39$_{-15.80}$}   &   \textcolor{red}{74.05$_{-13.38}$} 
\\
\hline
\hline
(+20,+20)   &  &     \textcolor{red}{83.59$_{-2.82}$}     &  \textcolor{red}{85.95$_{-2.39}$}     &  &    74.57$_{-3.42}$   &  77.78$_{-3.06}$    &  &  79.82$_{-2.30}$     &      82.63$_{-2.01}$     &  &     \textcolor{blue}{82.70$_{-2.49}$}  &  \textcolor{blue}{85.28$_{-2.15}$}  
\\
(+40,+40)   &  &   \textcolor{blue}{74.15$_{-12.26}$}   &   \textcolor{blue}{77.80$_{-10.54}$}   &  &   65.26$_{-12.73}$   &  69.43$_{-11.41}$  &  &   73.18$_{-8.94}$  &    76.80$_{-7.84}$    &  &    \textcolor{red}{76.69$_{-8.50}$}    &  \textcolor{red}{80.20$_{-7.23}$}      
\\
(+60,+60)   &  &   57.60$_{-28.81}$  &  62.88$_{-25.46}$   &  &  50.04$_{-27.95}$  &   55.42$_{-25.42}$  &  &   \textcolor{blue}{60.97$_{-21.15}$} &     \textcolor{blue}{65.93$_{-18.71}$}   &  &       \textcolor{red}{66.55$_{-18.64}$}   &  \textcolor{red}{71.44$_{-15.99}$} 
\\
(+80,+80)   &  &   41.30$_{-45.11}$  &   47.15$_{-41.19}$  &  &   34.27$_{-43.72}$  &   40.27$_{-40.57}$  &  &  \textcolor{blue}{45.37$_{-36.75}$}    &  \textcolor{blue}{51.47$_{-33.17}$}      &  &      \textcolor{red}{54.58$_{-30.61}$}   &  \textcolor{red}{60.74$_{-26.69}$} 
\\
(+100,+100) &  &    28.07$_{-58.34}$   &   33.73$_{-54.61}$    &  &   23.89$_{-54.10}$   &   29.60$_{-51.24}$  &  &  \textcolor{blue}{31.86$_{-50.26}$}      &  \textcolor{blue}{38.23$_{-46.41}$}  &  &    \textcolor{red}{43.36$_{-41.83}$}  &  \textcolor{red}{50.16$_{-37.27}$}
\\ 
\hline
\hline
(-20,-20)   &  &    \textcolor{blue}{84.35$_{-2.06}$}  &  \textcolor{blue}{86.62$_{-1.72}$}    &  &  76.40$_{-1.59}$   &   79.42$_{-1.42}$   &  &     80.71$_{-1.41}$     &   83.45$_{-1.19}$   &  &     \textcolor{red}{84.39$_{-0.80}$}    &  \textcolor{red}{86.76$_{-0.67}$} 
\\
(-40,-40)   &  &      \textcolor{blue}{78.10$_{-8.31}$}  &   \textcolor{blue}{81.24$_{-7.10}$}   &  &   70.27$_{-7.72}$   &  74.03$_{-6.81}$  &  &     76.51$_{-5.61}$    &     79.80$_{-4.84}$   &  &    \textcolor{red}{81.75$_{-3.44}$}    &   \textcolor{red}{84.55$_{-2.88}$}   
\\
(-60,-60)   &  &    67.73$_{-18.68}$  &  71.97$_{-16.37}$  &  &    59.56$_{-18.43}$  &   64.34$_{-16.50}$  &  &   \textcolor{blue}{68.73$_{-13.39}$}    &      \textcolor{blue}{72.93$_{-11.71}$}   &  &   \textcolor{red}{76.94$_{-8.25}$}  &    \textcolor{red}{80.46$_{-6.97}$}   
\\
(-80,-80)   &  &    54.60$_{-31.81}$    &     59.80$_{-28.54}$    &  & 46.60$_{-31.39}$   &  52.24$_{-28.60}$   &  &  \textcolor{blue}{57.54$_{-24.58}$}  &   \textcolor{blue}{62.84$_{-21.80}$}    &  &     \textcolor{red}{70.07$_{-15.12}$}   & \textcolor{red}{74.50$_{-12.93}$}
\\
(-100,-100) &  &   42.45$_{-43.96}$    &  48.10$_{-40.24}$   &  &    35.28$_{-42.71}$  &   41.35$_{-39.46}$   &  &   \textcolor{blue}{45.32$_{-36.80}$}  &    \textcolor{blue}{51.42$_{-33.22}$}   &  &      \textcolor{red}{61.54$_{-23.65}$}   & \textcolor{red}{66.92$_{-20.51}$}
\\ 
\hline
\hline
(+20,-20)   &  &  \textcolor{blue}{84.23$_{-2.18}$}    &    \textcolor{blue}{86.55$_{-1.79}$}  &  &     76.34$_{-1.65}$  &    79.37$_{-1.47}$   &  &   80.94$_{-1.18}$      &    83.62$_{-1.02}$    &  &     \textcolor{red}{84.32$_{-0.87}$}   &  \textcolor{red}{86.72$_{-0.71}$}
\\
(+40,-40)   &  &   \textcolor{blue}{77.90$_{-8.51}$}  &   \textcolor{blue}{81.09$_{-7.25}$}    &  &   70.36$_{-7.63}$  &   74.10$_{-6.74}$ &  &       76.58$_{-5.54}$   &      79.88$_{-4.76}$   &  &       \textcolor{red}{81.62$_{-3.57}$}       &   \textcolor{red}{84.46$_{-2.97}$}      
\\
(+60,-60)   &  &    67.29$_{-19.12}$  &    71.62$_{-16.72}$   &  & 59.61$_{-18.38}$    &   64.42$_{-16.42}$   &  &   \textcolor{blue}{68.63$_{-13.49}$}        &     \textcolor{blue}{72.90$_{-11.74}$}    &&        \textcolor{red}{76.80$_{-8.39}$}         &  \textcolor{red}{80.38$_{-7.05}$}         
\\
(+80,-80)   &  &     53.90$_{-32.51}$  &  59.21$_{-29.13}$   &  &   46.24$_{-31.75}$   &   51.98$_{-28.86}$   &  &    \textcolor{blue}{57.42$_{-24.70}$}    &     \textcolor{blue}{62.75$_{-21.89}$}   &  &     \textcolor{red}{69.76$_{-15.43}$}     &     \textcolor{red}{74.27$_{-13.16}$}   
\\
(+100,-100) &  &   41.69$_{-44.72}$   &   47.41$_{-40.93}$ &  &    34.90$_{-43.09}$   &   41.02$_{-39.82}$  &  &    \textcolor{blue}{45.40$_{-36.72}$}   &   \textcolor{blue}{51.44$_{-33.20}$}         &  &     \textcolor{red}{61.36$_{-23.83}$}  & \textcolor{red}{66.80$_{-20.63}$} 
\\ 
\hline
\hline
(-20,+20)   &  &   \textcolor{red}{83.46$_{-2.95}$}  &   \textcolor{red}{85.85$_{-2.49}$} &  &   74.74$_{-3.25}$  &    77.92$_{-2.92}$   &  &  79.75$_{-2.37}$   &    82.58$_{-2.06}$     &  &       \textcolor{blue}{82.95$_{-2.24}$}     &   \textcolor{blue}{85.49$_{-1.94}$} 
\\
(-40,+40)   &  &      \textcolor{blue}{74.47$_{-11.94}$}    &    \textcolor{blue}{78.05$_{-10.29}$}  &  &    65.13$_{-12.86}$    &   69.32$_{-11.52}$   &  &   73.13$_{-8.99}$    & 76.77$_{-7.87}$   &  &      \textcolor{red}{77.00$_{-8.19}$}   &  \textcolor{red}{80.46$_{-6.97}$} 
\\
(-60,+60)   &  &    58.05$_{-28.36}$   &   63.27$_{-25.07}$    &  &    50.19$_{-27.80}$   &  55.56$_{-25.28}$   &  &   \textcolor{blue}{61.13$_{-20.99}$}     &    \textcolor{blue}{66.03$_{-18.61}$} &  &      \textcolor{red}{66.87$_{-18.32}$}  &   \textcolor{red}{71.71$_{-15.72}$}  
\\
(-80,+80)   &  &     41.69$_{-44.72}$  &  47.52$_{-40.82}$    &  &    34.70$_{-43.29}$     &  40.69$_{-40.15}$     &  &   \textcolor{blue}{45.55$_{-36.57}$}      &  \textcolor{blue}{51.69$_{-32.95}$}   &  &    \textcolor{red}{54.96$_{-30.23}$} &  \textcolor{red}{61.06$_{-26.37}$}  
\\
(-100,+100) &  &   27.93$_{-58.48}$  &   33.67$_{-54.67}$    &  &  23.79$_{-54.20}$   &    29.60$_{-51.24}$    &  &   \textcolor{blue}{32.13$_{-49.99}$}     &  \textcolor{blue}{38.54$_{-46.10}$}  &  &    \textcolor{red}{43.67$_{-41.52}$}  &  \textcolor{red}{50.48$_{-36.95}$} 
\\ 
\hline
\end{tabular}
}
\end{table*}

\begin{table*}[!t]
\centering
\caption{\cq{ablation study on DPS with various segmentation points.}}
\label{DPS ABC}
\resizebox{1.0\linewidth}{!}{
\begin{tabular}{lcccccccccccccc}
\toprule
Padding   &   \multicolumn{2}{c}{LPN(ResNet)}  & &  \multicolumn{2}{c}{DPS-TL~(top-left,$\Delta H=|4|$)} &  & \multicolumn{2}{c}{DPS-BR~(bottom-right,$\Delta H=|4|$)} & & \multicolumn{2}{c}{DPS(center)} 
   && \multicolumn{2}{c}{SDPL(ResNet)} 
\\ \cline{2-3} \cline{5-6} \cline{8-9}  \cline{11-12} \cline{14-15}
pixel(P)   & Recall@1        & AP   && Recall@1        & AP     && Recall@1        & AP    && Recall@1        & AP  &&  Recall@1        & AP
\\ \midrule
(0,0)       &  77.99$_{-0}$ & 80.84$_{-0}$ &&  81.72$_{-0}$      &     84.45$_{-0}$   & &     81.12$_{-0}$   &      83.88$_{-0}$  &&  \textcolor{blue}{85.08$_{-0}$}  &  \textcolor{blue}{87.31$_{-0}$}  &&       \textcolor{red}{85.19$_{-0}$}  & \textcolor{red}{87.43$_{-0}$} 
\\ \hdashline
(+100,0)    &  47.73$_{-30.26}$ &  53.35$_{-27.49}$  &&    \textcolor{blue}{66.91$_{-14.81}$}             &    \textcolor{blue}{71.71$_{-12.74}$}       &&    65.48$_{-15.64}$    &     70.31$_{-13.57}$    &&    66.50$_{-18.58}$   & 71.37$_{-15.94}$   &&  \textcolor{red}{69.39$_{-15.80}$}   &   \textcolor{red}{74.05$_{-13.38}$}
\\
(+100,+100) &     23.89$_{-54.10}$   &  29.60$_{-51.24}$   &&  38.73$_{-42.99}$      &     45.56$_{-38.89}$    &&      \textcolor{red}{45.61$_{-35.51}$}      &    \textcolor{red}{51.81$_{-32.07}$}
&&    42.83$_{-42.25}$  &  49.44$_{-37.87}$    &&  \textcolor{blue}{43.36$_{-41.83}$}   &    \textcolor{blue}{50.16$_{-37.27}$}
\\
(-100,-100) &   35.28$_{-42.71}$  &   41.35$_{-39.46}$  &&  56.79$_{-24.93}$     &   62.39$_{-22.06}$       &&      50.64$_{-30.48}$           &     56.74$_{-27.14}$    &&   \textcolor{blue}{57.40$_{-27.68}$}   &   \textcolor{blue}{63.07$_{-24.24}$}   && \textcolor{red}{61.54$_{-23.65}$}  &   \textcolor{red}{66.92$_{-20.51}$}
\\
(+100,-100) &   34.90$_{-43.09}$  & 41.02$_{-39.82}$  && 56.80$_{-24.92}$     &   62.49$_{-21.96}$     &&    50.63$_{-30.49}$    &  56.72$_{-27.16}$     &&  \textcolor{blue}{57.40$_{-27.68}$}  &  \textcolor{blue}{63.12$_{-24.19}$}  &&
 \textcolor{red}{61.36$_{-23.83}$} &   \textcolor{red}{66.80$_{-20.63}$} 
\\
(-100,100)  &    32.13$_{-49.99}$   &    38.54$_{-46.10}$  &&   38.76$_{-42.96}$    &    45.62$_{-38.83}$    &&     \textcolor{red}{45.39$_{-35.73}$}    &     \textcolor{red}{51.74$_{-32.14}$}  &&   43.37$_{-41.71}$  &  49.92$_{-37.39}$  &&
\textcolor{blue}{43.67$_{-41.52}$}  &   \textcolor{blue}{50.48$_{-36.95}$}
\\
\bottomrule
\end{tabular}
}
\end{table*}

\cq{\textbf{Source of SDPL's anti-offset property.}}
\cq{We hold the opinion that both DPS and SFS contribute to the anti-offset property of SDPL.
Compared with LPN, our DPS retains global features while generating a variety of fine-grained features, thus achieving superior retrieval performance and stronger anti-offset capacity in different scenarios, as shown in columns 1 and 4 of Table~\ref{DPS ABC}.}
\cq{To further explore the impact of segmentation points on DPS, we adjust the segmentation point of DPS according to Eq.~\ref{segmentation point adjustment}, and test the retrained models under various deviation conditions.
As shown in Table~\ref{DPS ABC}, DPS~(center) shows the best performance in the default scenarios, confirming that most of the retrieval targets in University1652~\cite{zheng2020university} are centered.
When the query images are randomly shifted (\textit{i.e.}, non-centered scenarios), by comparing DPS, DPS-TL~(top-left), and DPS-BR~(bottom-right), it can be observed that the performance degradation of DPS with different segmentation points is direction-dependent.
For instance, as shown in Table~\ref{DPS ABC}, DPS-BR shows a better anti-offset property when the query images are offset to the bottom-right~($+P$,$+P$), whereas DPS-TL shows less degradation when the query images are offset to the top-left~($-P$,$-P$).
This phenomenon inspires SFS, fusing multiple sets of DPS to further improve the anti-offset property of our solution.}
\cq{To further show the effectiveness of SFS, we apply the SFS to LPN and DPS respectively, with results as shown in Tables~\ref{padding results LPN and SDPL} and \ref{DPS ABC}.
When SFS is injected into LPN, we observe that the new model, namely LPN+SFS, shows stronger retrieval performance and less performance degradation than LPN in various scenarios, which demonstrates the feasibility and validity of SFS.
Compared with DPS, our SPDL~(DPS+SFS) shows superior robustness to position deviations in offset scenarios of five directions, as shown in columns 4 and 5 of Table~\ref{DPS ABC}.}

\begin{figure}[!t]
    \centering
    \includegraphics[width=0.85\linewidth]{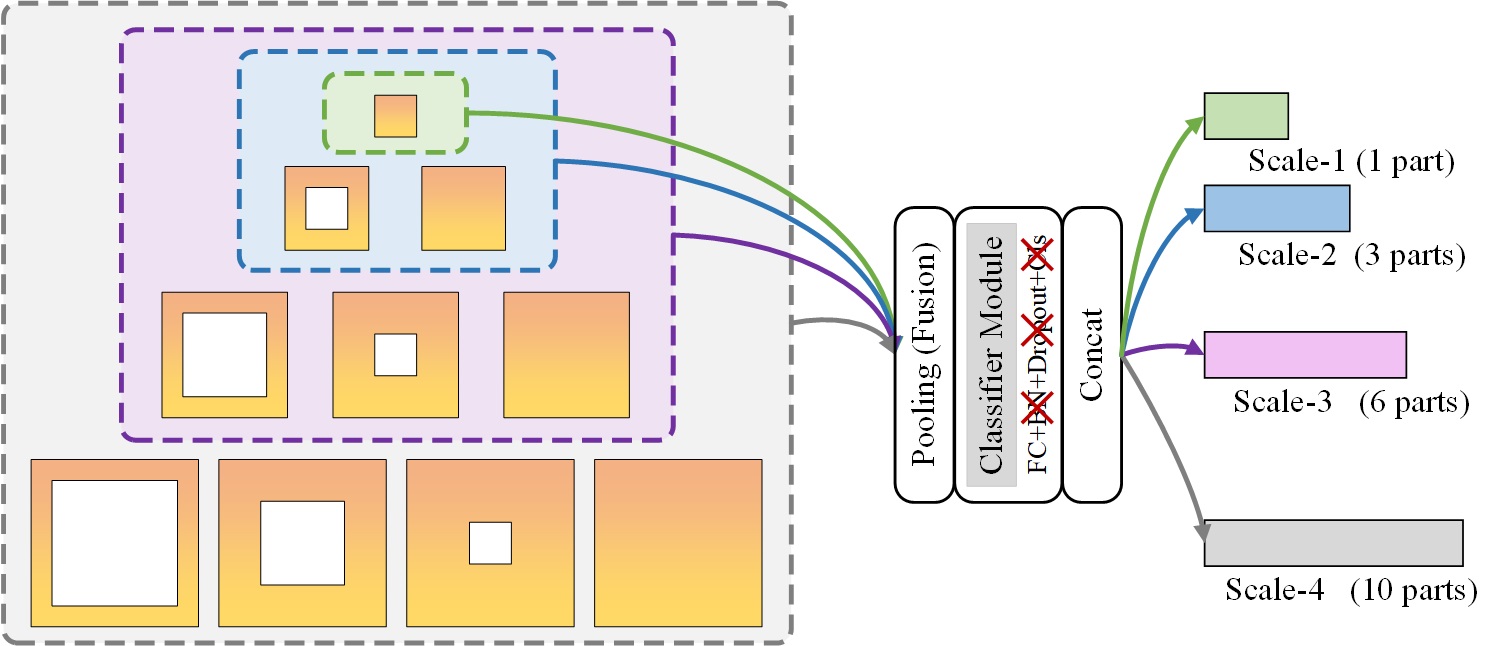}
    \caption{Simplified diagram of various combination of parts. Depending on distance to center, we sequentially adopt more descriptors generated by parts for inference. As scale increases, more contextual information is used. Note that during test phase, classifier module's BN, Dropout, and Cls are not used.}
    \label{parts combined diagram}
\end{figure}

\begin{figure}[!t]\color{red}
    \centering
    \includegraphics[width=1.0\linewidth]{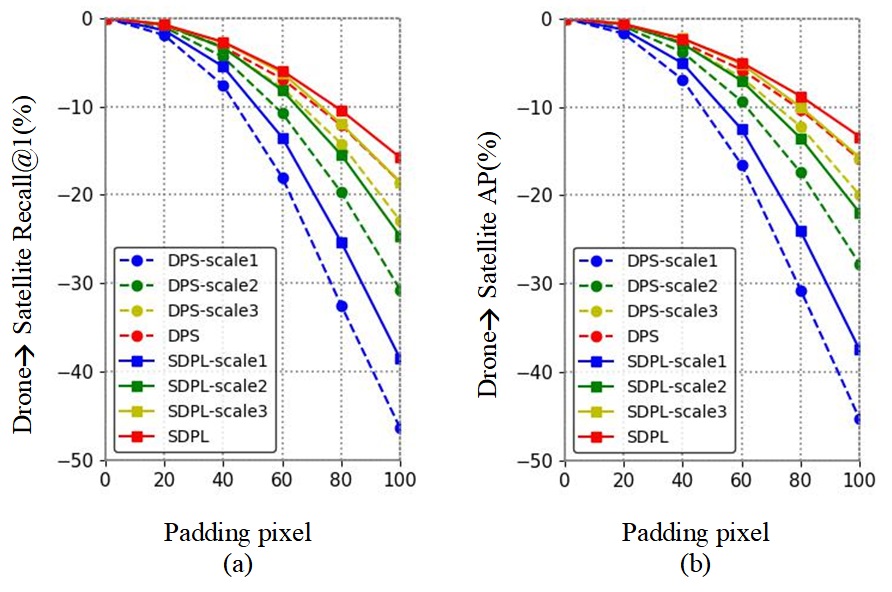}
    \caption{Impact of number of 1-D pads on AP and Recall@1. Vertical axis: magnitude of decrease in accuracy.}
    \label{multiscale padding}
\end{figure}

\cq{\textbf{Does SDPL leverage contextual information?}
DPS divides high-level features into a set of parts, each containing various content.
In practice, visual descriptors corresponding to all parts are concatenated, and their similarity is calculated based on Euclidean distance.
Based on the observation that high-level features and RGB features have similar distributions~(see Fig.~\ref{heatmap_visual}), we design various combinations of different parts according to the distance to the feature center, thus proving that SDPL exploits contextual information.
As shown in Fig.~\ref{parts combined diagram}, we sequentially adopt more parts for inference and classify the combinations into four classes, \textit{i.e.}, scale-1,-2, -3, -4.
In short, as the scale increases, the resolution of parts gradually increases, meaning that more contextual information is used for inference.
Table~\ref{various combination of parts} shows the experimental results of DPS and SDPL respectively. For DPS and SDPL, the retrieval accuracy of scale-3 and scale-4 is significantly higher than that of scale-1 and scale-2.
As shown in Figure~\ref{multiscale padding}, DPS and SDPL suffer less performance degradation in complex scenarios as the scale increases.
This indicates that contextual information~(the surrounding environment) is crucial to the retrieval accuracy and anti-offset ability of SPDL.
Due to the adjustment of the partition center, SDPL of scale-3 sees almost all the contextual feature~(see Fig.~\ref{shifting strategy img}(a)), resulting in stable accuracy in the default scenario.
However, we can discover that the new parts can further promote   SDPL to endure less performance degradation than the scale-3 model~(red and yellow lines, Fig.~\ref{multiscale padding}).}

\begin{table}[!t]
\centering
\caption{\cq{Ablation study of using various combination of different parts during inference.}}
\label{various combination of parts}
\resizebox{1.0\linewidth}{!}{
\begin{tabular}{lccccccccccc}
\hline
Padding   & \multicolumn{2}{c}{scale-1} &  & \multicolumn{2}{c}{scale-2} &  & \multicolumn{2}{c}{scale-3} &  & \multicolumn{2}{c}{DPS~(scale-4)} 
\\ \cline{2-3} \cline{5-6} \cline{8-9} \cline{11-12} 
pixel (P) & Recall@1       & AP         &  & Recall@1       & AP         &  & Recall@1       & AP         &  & Recall@1      & AP    
\\ \hline
(0,0)     & 67.60          & 71.41      &  & 79.70          & 82.39      &  & \textcolor{blue}{84.33}          & \textcolor{blue}{86.56}      &  &  \textcolor{red}{85.08}         &  \textcolor{red}{87.31}   
\\
(+20,0)   & 65.65          & 69.68      &  & 78.66          & 81.49       &  & \textcolor{blue}{83.49}           & \textcolor{blue}{85.86}      &  &  \textcolor{red}{84.26}         &  \textcolor{red}{86.61}   
\\
(+40,0)   & 59.94          & 64.48      &  & 75.26          & 78.55      &  & \textcolor{blue}{80.91}          & \textcolor{blue}{83.67}      &  &  \textcolor{red}{81.79}          &  \textcolor{red}{84.56}   
\\
(+60,0)   & 49.58          & 54.87      &  & 68.92          & 73.02      &  & \textcolor{blue}{76.42}          & \textcolor{blue}{79.86}      &  &  \textcolor{red}{78.19}         &  \textcolor{red}{81.50}  
\\
(+80,0)   & 35.11             & 40.71      &  & 59.98          & 64.97      &  & \textcolor{blue}{71.11}          & \textcolor{blue}{74.34}      &  &  \textcolor{red}{72.93}         &  \textcolor{red}{77.00}  
\\
(+100,0)  & 21.22         & 26.13     &  & 48.96          & 54.63      &  & \textcolor{blue}{61.39}          & \textcolor{blue}{66.59}      &  &  \textcolor{red}{66.50}         &  \textcolor{red}{71.37}   
\\ \hline
\hline
Padding   & \multicolumn{2}{c}{scale-1} &  & \multicolumn{2}{c}{scale-2} &  & \multicolumn{2}{c}{scale-3} &  & \multicolumn{2}{c}{SDPL~(scale-4)} \\ \cline{2-3} \cline{5-6} \cline{8-9} \cline{11-12} 
pixel (P) & Recall@1       & AP         &  & Recall@1       & AP         &  & Recall@1       & AP         &  & Recall@1      & AP       \\ \cline{1-9} \cline{11-12} 
(0,0)     & 63.24          & 67.41      &  & 77.35          & 80.34      &  & \textcolor{red}{85.52}          & \textcolor{red}{87.59}      &  & \textcolor{blue}{85.19}        & \textcolor{blue}{87.43}    \\
(+20,0)   & 61.83          & 66.14      &  & 76.61          & 79.72      &  & \textcolor{red}{84.73}          & \textcolor{red}{86.94}      &  & \textcolor{blue}{84.42}         & \textcolor{blue}{86.78}    \\
(+40,0)   & 57.72          & 62.33      &  & 73.98          & 77.46      &  & \textcolor{red}{82.77}          & \textcolor{red}{85.32}      &  & \textcolor{blue}{82.46}         & \textcolor{blue}{85.13}    \\
(+60,0)   & 49.72          & 54.87      &  & 69.17          & 73.28      &  & \textcolor{red}{79.29}          & \textcolor{blue}{82.42}      &  & \textcolor{blue}{79.22}         & \textcolor{red}{82.43}    \\
(+80,0)   & 37.90          & 43.36      &  & 61.87          & 66.80      &  & \textcolor{blue}{73.53}         & \textcolor{blue}{77.56}      &  & \textcolor{red}{74.73}         & \textcolor{red}{78.63}    \\
(+100,0)  & 24.71          & 29.96      &  & 52.67          & 58.37      &  & \textcolor{blue}{66.98}          & \textcolor{blue}{71.86}      &  & \textcolor{red}{69.37}         & \textcolor{red}{74.05}    \\ \hline
\end{tabular}
}
\end{table}

\textbf{Shifting values.} 
To explore the effect of offset parameters $\{\Delta H_{1},\Delta H_{2}\}$~($\Delta H_{1,2}$ for convenience) on the anti-offset property of SPDL, we conduct experiments changing only the offset values.
As shown in Table~\ref{experiments about Shifting values}, when $\Delta H_{1,2}$=$|0|or|1|$, the model achieves the best Recall@1 and AP in the case of a small padding pixel~($\text{P}\in[0,40]$) because there are $3\times$ partitions to thoroughly mine central features.
However, as the offset increases, the performance degradation of the model is significant; for example, when P=100, Recall@1 of model with $\Delta H_{1,2}$=$|0|$ drops by 18.94\%.
On the contrary, when the offset $\Delta H_{1,2}$=$|3|or|4|$, the model can explicitly fit the offset scenarios.
Although initial performance of the model is weak, its performance decreases slowly with the increase of padding pixels, in short, excellent anti-offset ability. 
For example, when $\Delta H_{1,2}$=$|3|$ and P=100, the Recall@1 degrades by 14.79\%, raising it by 4.15\% over the model with $\Delta H_{1,2}$=$|0|$.
To balance model performance and anti-offset ability, $\Delta H_{1,2}$=$|2|$ is selected.

\begin{table*}[!t]
\centering
\caption{\cq{Performance comparison of SDPL framework with different shifting values $\{\Delta H_{1},\Delta H_{2}\}$.}}
\label{experiments about Shifting values}
\resizebox{1.0\linewidth}{!}{
\begin{tabular}{llcclcclcclcclcc}
\toprule
Padding  && \multicolumn{2}{c}{$\Delta H_{1,2}=|0|$}   &&\multicolumn{2}{c}{$\Delta H_{1,2}=|1|$}   && \multicolumn{2}{c}{$\Delta H_{1,2}=|2|$}   && \multicolumn{2}{c}{$\Delta H_{1,2}=|3|$}  &&
 \multicolumn{2}{c}{$\Delta H_{1,2}=|4|$}    \\ \cline{3-4} \cline{6-7} \cline{9-10} \cline{12-13} \cline{15-16} 
Pixel($P$)  && \multicolumn{1}{c}{Recall@1} & \multicolumn{1}{c}{AP} &  & \multicolumn{1}{c}{Recall@1} & \multicolumn{1}{c}{AP} &  & \multicolumn{1}{c}{Recall@1} & \multicolumn{1}{c}{AP} &  & \multicolumn{1}{c}{Recall@1} & \multicolumn{1}{c}{AP} &  & \multicolumn{1}{c}{Recall@1} & \multicolumn{1}{c}{AP}    \\ 
\midrule
(0,0)    && \textcolor{red}{85.88} & \textcolor{red}{88.00}    && \textcolor{blue}{85.56} & \textcolor{blue}{87.76}    && {85.19}  & {87.43}  && 83.63      & 86.08    && 81.93    & 84.65 \\ 
\hdashline
(+20,0)  &&  \textcolor{red}{84.92$_{-0.96}$} & \textcolor{red}{87.20$_{-0.80}$}    &&  \textcolor{blue}{84.58$_{-0.98}$}   & \textcolor{blue}{86.94$_{-0.82}$}   && {84.42$_{-0.77}$}    & {86.78$_{-0.65}$}   && 82.92$_{-0.71}$   & 85.47$_{-0.61}$   && 81.12$_{-0.81}$   & 83.98$_{-0.67}$            
\\
(+40,0)  &&  \textcolor{blue}{82.50$_{-3.38}$} & \textcolor{blue}{85.17$_{-2.83}$}    && \textcolor{red}{82.56$_{-3.00}$}    & \textcolor{red}{85.24$_{-2.52}$}     && {82.46$_{-2.73}$}   & {85.13$_{-2.30}$}    && 80.95$_{-2.68}$  & 83.80$_{-2.28}$   && 79.34$_{-2.59}$     & 82.48$_{-2.17}$      
\\
(+60,0)  && {78.92$_{-6.96}$} & {82.15$_{-5.85}$}    && \textcolor{blue}{78.95$_{-6.61}$}    & \textcolor{blue}{82.20$_{-5.56}$}     && \textcolor{red}{79.22$_{-5.97}$}  & \textcolor{red}{82.43$_{-5.00}$}  && 78.09$_{-5.54}$    & 81.39$_{-4.69}$  && 76.72$_{-5.21}$    & 80.23$_{-4.42}$     
\\
(+80,0)  && {73.50$_{-12.38}$} & {77.55$_{-10.45}$}    &&  73.88$_{-11.68}$    & 77.87$_{-9.89}$    && \textcolor{red}{74.73$_{-10.46}$}    & \textcolor{red}{78.63$_{-8.80}$}    && \textcolor{blue}{74.13$_{-9.50}$}   & \textcolor{blue}{78.01$_{-8.07}$}   && 72.51$_{-9.42}$    & 76.63$_{-8.02}$               
\\
(+100,0) && {66.94$_{-18.94}$} & {71.83$_{-16.17}$}    && 67.61$_{-17.95}$        & 72.45$_{-15.31}$    && \textcolor{red}{69.39$_{-15.80}$}   & \textcolor{red}{74.05$_{-13.38}$}     && \textcolor{blue}{68.84$_{-14.79}$}     & \textcolor{blue}{73.47$_{-12.61}$}    && 67.54$_{-14.39}$  & 72.29$_{-12.36}$     \\ 
\bottomrule
\end{tabular}
}
\end{table*}

\begin{table}[!t]
\centering
\caption{\cq{Performance comparison of SDPL with different number of parts. $N_{SPS}$ denotes the number of basic parts, and $N_{DPS}$ denotes the number of parts generated by DPS.}}
\label{partition number}
\resizebox{1.0\linewidth}{!}{
\begin{tabular}{ccccccc}
\toprule
Numbers && \multicolumn{2}{c}{Drone$\rightarrow$Satellite} && \multicolumn{2}{c}{Satellite$\rightarrow$Drone} \\ \cline{3-4} \cline{6-7} 
$N_{SPS}\rightarrow N_{DPS}$ && Recall@1  & AP    && Recall@1   & AP    \\ \midrule
1 $\rightarrow$ 1   &&  74.52   &   78.23      &&  83.74    &  73.02       \\
2 $\rightarrow$ 3  &&   81.25   &   84.04      &&  89.16    &  79.46      \\
3 $\rightarrow$ 6   &&  \textcolor{blue}{84.71}   &   \textcolor{blue}{86.98}      &&  \textcolor{blue}{88.87}    &  \textcolor{blue}{82.61}       \\
4 $\rightarrow$ 10 &&   \textcolor{red}{85.19}   &   \textcolor{red}{87.43}      &&  \textcolor{red}{89.30}    &  \textcolor{red}{82.75}       \\ \bottomrule
\end{tabular}
}
\end{table}

\begin{figure}[!t]
    \centering
    \includegraphics[width=1.0\linewidth]{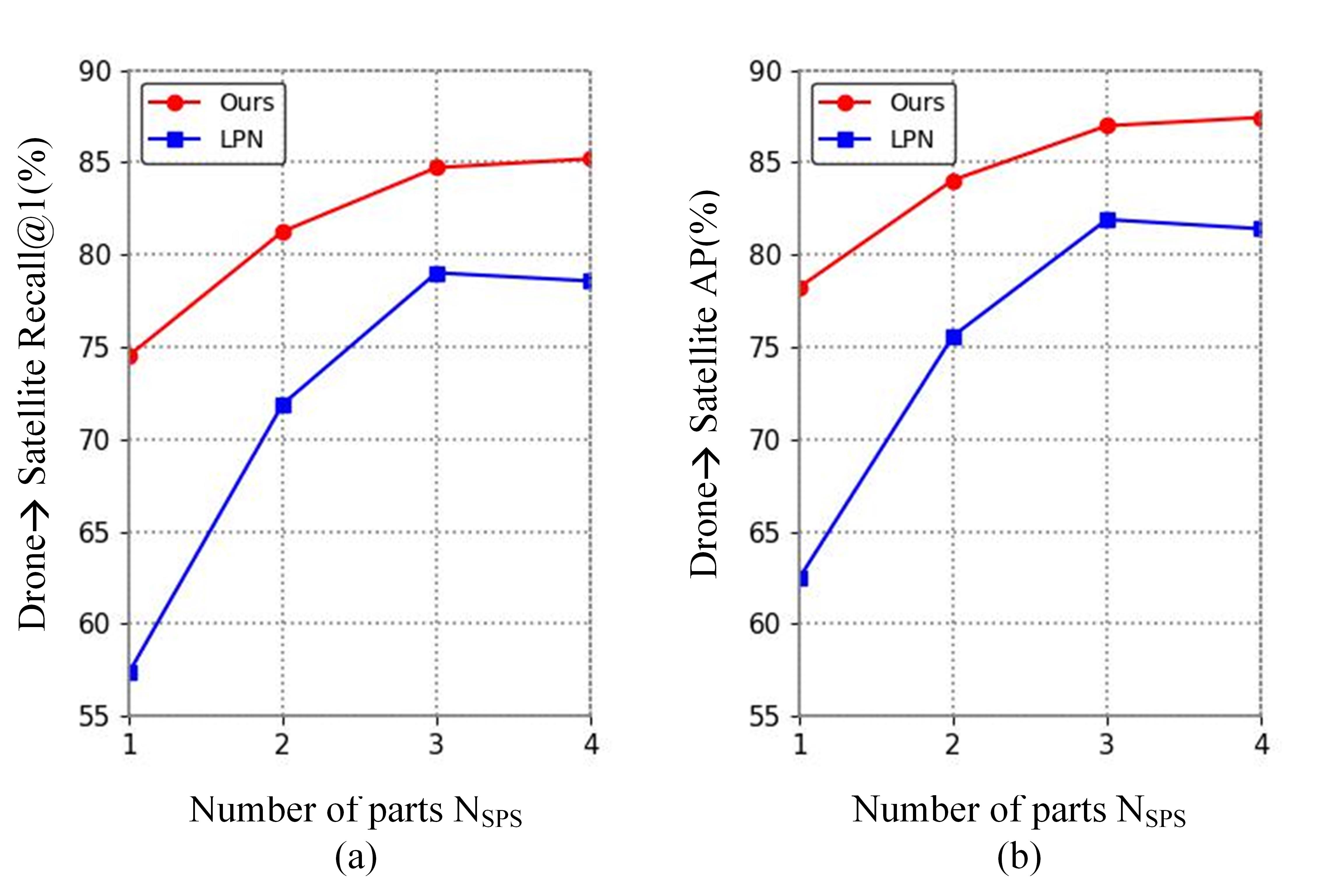}
    \caption{Comparison of LPN and SPDL with different numbers of parts.}
    \label{compare lpn and denselpn in N}
\end{figure}

\textbf{Number of partitions}. 
The number of parts $N_{DPS}$ is a key hyper-parameter in our network. 
Following Eq.~\ref{NDPS AND NSPS}, the increment of $N_{DPS}$ is nonlinear.
When $N_{DPS}=1$, the model is degraded into our baseline.
Considering the suitability of basic part size with feature shifting strategy and computational costs of the network, we specify that $Max\left[N_{SPS}\right]=4$.
As shown in Table~\ref{partition number}, for both tasks, \textit{i.e.}, (Drone$\rightarrow$Satellite) and (Satellite$\rightarrow$Drone), with the increment of $N_{DPS}$, we observe that both the Recall@1 and AP values improve significantly.
Moreover, as shown in Fig.~\ref{compare lpn and denselpn in N}, regardless of $N_{SPS}$, SPDL can perform better than LPN~\cite{wang2021each}.
Concatenating more contextual information parts can improve the discriminability of the final feature descriptor, and our dense partition strategy always contains a global feature to constrain the feature consistency of the cross-view~\cite{zhuang2021faster,zhu2023uav}.

\begin{table}[!t]
\centering
\caption{\cq{Performance comparison of SDPL with different input sizes on University-1652.}}
\label{Image size}
\resizebox{1.0\linewidth}{!}{
\begin{tabular}{ccccccc}
\toprule
\multirow{2}{*}{Image size} &  & \multicolumn{2}{c}{Drone$\rightarrow$Satellite} &  & \multicolumn{2}{c}{Satellite$\rightarrow$Drone} \\ \cline{3-4} \cline{6-7} 
&  & Recall@1   & AP       &  & Recall@1   & AP       \\ \midrule
224$\times$224    &&    74.90   &   78.10      &&   85.16     &  74.71     \\
256$\times$256    &&    79.74   &   82.50      &&   85.45     &  77.74     \\
320$\times$320    &&    82.75   &   85.29      &&   88.16     &  81.23     \\
384$\times$384    &&    \textcolor{blue}{83.27}   &  \textcolor{blue}{85.74}      &&   \textcolor{blue}{89.30}    &  \textcolor{blue}{81.69}     \\
512$\times$512    &&    \textcolor{red}{85.19}   &   \textcolor{red}{87.43}      &&   \textcolor{red}{89.30}     &  \textcolor{red}{82.75}     \\ \bottomrule
\end{tabular}
}
\end{table}

\begin{figure}[!t]
    \centering
    \includegraphics[width=1.0\linewidth]{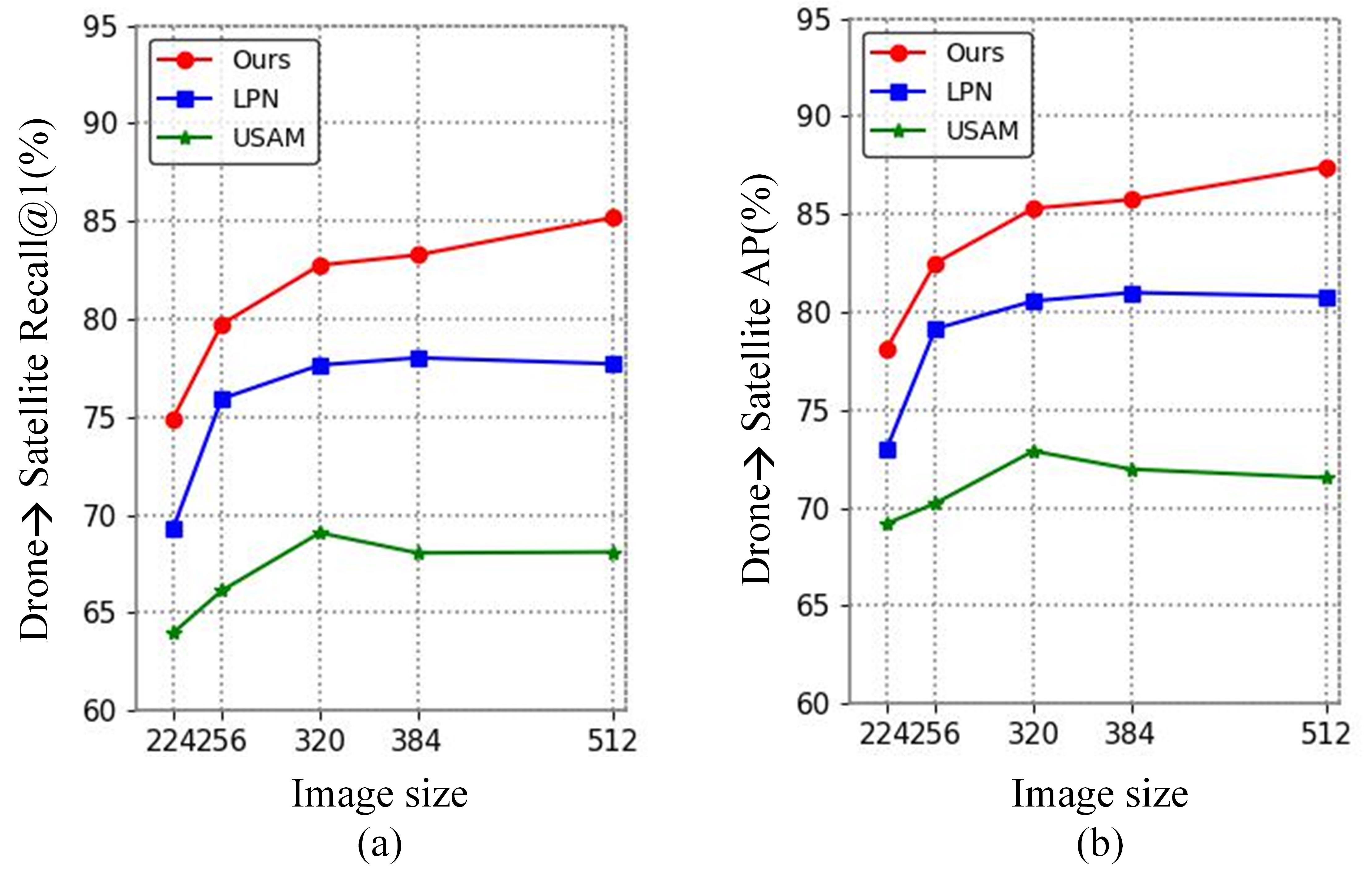}
    \caption{\cq{Impact of different input sizes on Recall@1 and AP.}}
    \label{Image size visual}
\end{figure}

\textbf{Input image size}. 
As shown in Table~\ref{Image size}, when the input image resolution changed from 224 to 512, model performance improved continuously.
This shows that SDPL could adapt to resolution input, \textit{i.e.}, it could be applied to a wide range of real-world scenarios.  
The performance of SDPL is compared with that of the existing model at different resolutions, as shown in Figure~\ref{Image size}.  
The performance of 
LPN~\cite{wang2021each}  was visibly degraded when the resolution was reduced to 224. 
With the resolution increasing from 256 to 512, SDPL achieves the most significant performance improvement over LPN and USAM, because the proposed dense partition strategy mines richer semantic features.
In general, our method is more competitive in application scenarios with time and space complexity constraints.

\section{Conclusions}\label{Conclusions}
In this paper, we present simple and effective representation learning for UAV-view geo-localization.
To explore contextual information being robust to position deviations in a unified architecture, we propose SDPL, which included dense partition and shifting-fusion strategies.
The dense partition strategy divides features into shape-diverse parts to mine fine-grained representations and maintain global structure, thus mitigating the impact of position shifting.
\cq{We focus on the degradation caused by position deviations, and propose a shifting-fusion strategy, which generates multiple sets of partitions based on various segmentation points to cope with non-centered scenarios, followed by a weight estimation module for adaptive partition fusion.
In experiments, SDPL achieves competitive results on two benchmarks: University-1652~\cite{zheng2020university} and SEUS-200~\cite{zhu2023sues}.
Ablation experiments show that SDPL has superior robustness for target non-centered scenarios.
Besides, our SDPL reveals satisfactory compatibility with a variety of backbone networks (\textit{e.g.}, ResNet and Swin).}
In the future, we plan to extend our approach to other cross-view matching scenarios, such as ground- and satellite-view.


 

%

\bibliographystyle{IEEEtran}
\bibliography{ref.bib}

\vfill

\end{document}